\documentclass[10pt,twocolumn,letterpaper]{article}

\usepackage{cvpr}
\usepackage{times}
\usepackage{epsfig}
\usepackage{graphicx}
\usepackage{amsmath}
\usepackage{amssymb}
\usepackage{float}
\usepackage{booktabs}
\usepackage{multirow}
\usepackage[british]{babel}
\usepackage[autostyle]{csquotes}
\usepackage[T1]{fontenc}
\newcommand\tab[1][0.4cm]{\hspace*{#1}}
\usepackage{enumitem}
\usepackage[dvipsnames]{xcolor}
\usepackage{placeins}
\usepackage{makecell}
\usepackage{caption} 
\usepackage{appendix}

\newcommand{\red}[1]{\textcolor{red}{#1}}
\newcommand{\myit}[1]{#1}
\newcommand{\green}[1]{\textcolor{OliveGreen}{#1}}
\newcommand{\myparagraph}[1]{\vspace{1pt}\noindent{\textbf{#1.}}}
\newcommand{\vqacv}{{CV-VQA}}
\newcommand{\vqaiv}{{IV-VQA}}
\newcommand{\arr}{{$\rightarrow$}}

\newcommand{\RS}[1]{}
\newcommand{\VA}[1]{}
\newcommand{\mf}[1]{}
\newcommand{\mario}[1]{}

\usepackage[pagebackref=true,breaklinks=true,letterpaper=true,colorlinks,bookmarks=false]{hyperref}

\cvprfinalcopy %

\def\cvprPaperID{***} %

\newcommand{\authSpace}{\quad\quad}

\begin{document}

\title{Towards Causal VQA: Revealing and Reducing Spurious Correlations by Invariant and Covariant Semantic Editing}

\author{
  Vedika Agarwal\textsuperscript{1} \authSpace Rakshith Shetty\textsuperscript{2} \authSpace Mario Fritz\textsuperscript{3}
  \\[2mm]
  \textsuperscript{1}TomTom \authSpace
  \textsuperscript{2}Max Planck Institute for Informatics \authSpace \textsuperscript{3} CISPA Helmholtz Center for Information Security \\
  Saarland Informatics Campus\\
  \textsuperscript{1}{\tt\small vedika.agarwal@tomtom.com} \authSpace \textsuperscript{2}{\tt\small rakshith.shetty@mpi-inf.mpg.de} \authSpace \textsuperscript{3}{\tt\small fritz@cispa.saarland}
}
\maketitle

\begin{abstract}

 Despite significant success in Visual Question Answering (VQA), VQA models have been shown to be notoriously brittle to linguistic variations in the questions. Due to deficiencies in models and datasets, today's models often rely on correlations rather than predictions that are causal w.r.t. data. 
 In this paper, we propose a novel way to analyze and measure the robustness of the state of the art models w.r.t semantic visual variations as well as propose ways to make models more robust against spurious correlations. 
 Our method performs automated semantic image manipulations and tests for consistency in model predictions to quantify the model robustness as well as generate synthetic data to counter these problems. 
 We perform our analysis on three diverse, state of the art VQA models and diverse question types with a particular focus on challenging counting questions. In addition, we show that models can be made significantly more robust against inconsistent predictions using our edited data. Finally, we show that results also translate to real-world error cases of state of the art models, which results in improved overall performance.

\end{abstract}

\mario{we need to unify the appearance of the tables. every table looks different, I prefer the appearance of booktabs like I've editted this one \autoref{table:flipping_0.1_0.1}. Rakshith has some example how this can be done for subcolumns in his paper. Also there seems to be something wrong with the captions. In your submission they are too close to the tables - there should be a small space. please look into it.}

\section{Introduction}
\begin{figure}[t]
\centering
\small
  \begin{tabular}{l l l l l l}

    \multicolumn{5}{c}{\myit{Q}: Is this a kitchen? }\\
    \multicolumn{5}{l}{ \tab[1.7cm] \myit{A}: no    \tab[1.3cm]   \textit{toilet removed}; \myit{A}: no }\\
    \multicolumn{5}{c}{\includegraphics[ trim= 10 20 10 140, clip,height=0.28\linewidth,width=0.34\textwidth]{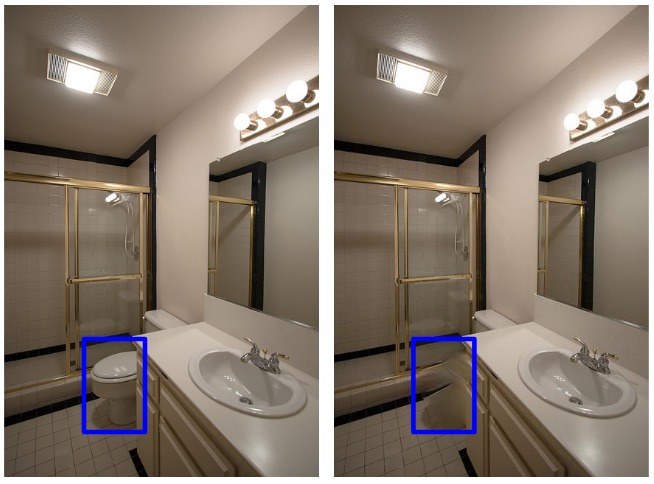}}\\
    & Baseline & Ours & Baseline &  Ours\\ 
    CL & \green{no} & \green{no} &  \red{yes} &\green{no} \\
    SAAA  & \green{no} & \green{no}&   \green{no} &\green{no}  \\ 
    SNMN &  \green{no} & \green{no} &  \red{yes} &\green{no}  \\

    \Xhline{2\arrayrulewidth}
    
    & \multicolumn{4}{c}{\myit{Q}: What color is the balloon? } \\
    \multicolumn{5}{l}{ \tab[1.8cm] \myit{A}: red    \tab[1.1cm]   \textit{umbrellas removed}; \myit{A}: red}\\

    \multicolumn{5}{c}{\tab \includegraphics[width=0.35\textwidth, trim= 0 0 0 0, clip]{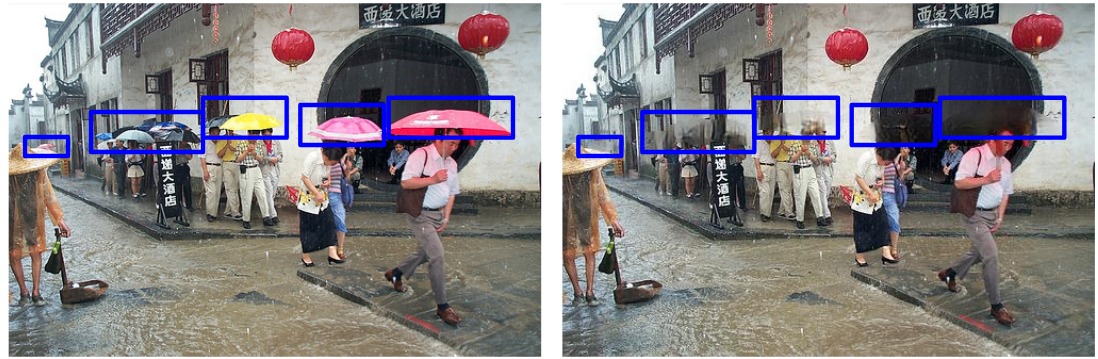}}\\
    & Baseline & Ours & Baseline & Ours\\ 
    CL & \red{pink} & \green{red}  & \green{red} &\green{red} \\
    SAAA  & \red{pink} & \green{red}  & \green{red} &\green{red}  \\ 
    SNMN &  \red{pink} & \green{red}  & \green{red} &\green{red}  \\
    
    \Xhline{2\arrayrulewidth}
    
    \multicolumn{5}{c}{\myit{Q}: How many zebras are there in the picture? } \\
    \multicolumn{5}{l}{\tab[1.8cm] \myit{A}: 2 \tab[1.3cm] \textit{zebra removed} \myit{A}: 1}\\
    \multicolumn{5}{c}{\includegraphics[width=0.34\textwidth, trim= 0 0 0 0, clip]{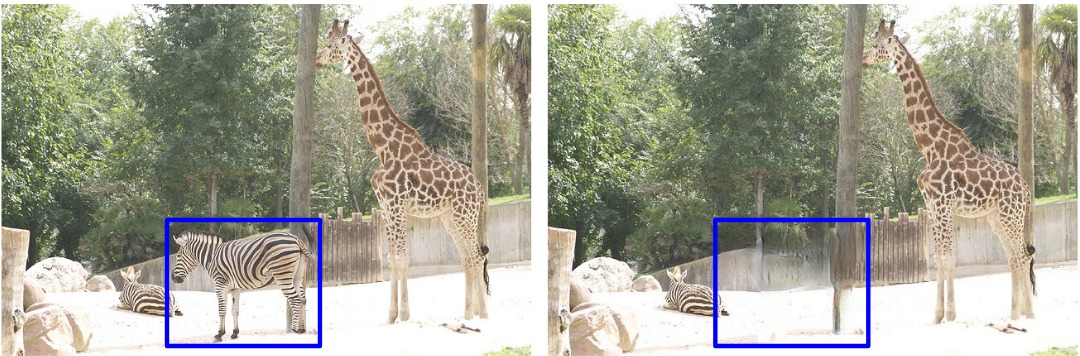}}\\
    & Baseline & Ours & Baseline & Ours\\ 
    CL & \green{2} & \green{2}  & \red{2} &\green{1} \\
    SAAA  & \green{2} & \green{2}  & \red{2} &\green{1}  \\ 
    SNMN &  \green{2} & \green{2}  & \red{2} &\green{1}  \\
  \end{tabular}
  \vspace*{-1mm}
  \caption{VQA models change their predictions as they exploit spurious correlations rather than causal relations based on the evidence. Shown above are predictions of 3 VQA models on original and synthetic images from our proposed \vqaiv{}  and \vqacv{} datasets. `Ours' denote the models robustified with our proposed data augmentation strategy.
  \RS{Order should change to 1,3,2. Low priority: figure alignments are off, to fix it images need be split and added in the right column} \mario{ideall make text "toilet removed; A:no" centered above image}}
  \vspace*{-4mm}
  \label{fig:teaser}
\end{figure}

VQA allows interaction between images and language, with diverse applications such as interacting with chat bots to assisting visually impaired people. In these applications we expect a model to answer truthfully and based on the evidence in the image and the actual intention of the question. Unfortunately, this is not always the case even for state of the art methods. Instead of ``sticking to the facts'', models frequently rely on spurious correlations and follow biases induced by data and/or model. 
For instance, recent works  \cite{marcus_paper, data_aug_sunny_dark_outside} have  shown that the VQA models are brittle to linguistic variations in questions/answers.  Shah \emph{et~al.} in \cite{marcus_paper} introduced VQA-Rephrasings dataset to expose the brittleness of the VQA models to linguistic variations and proposed cyclic consistency to improve their robustness. They show that if a model answers \textquote{Yes}  to the question: \textquote{Is it safe to turn left?}, it answers \textquote{No} when the question is rephrased to \textquote{Can one safely turn left?}. Similarly Ray \emph{et~al.} in  \cite{data_aug_sunny_dark_outside} introduced ConVQA to quantitatively evaluate the consistency for VQA towards different generated entailed questions and proposed data augmentation module to make the models more consistent.

While previous works have studied linguistic modifications, our contribution is the first systematic study of automatic visual content manipulations at scale. Analogous to rephrasing questions for VQA, images can also be semantically edited to create different variants where the same question-answer~(QA) pair holds. One sub-task of this broader semantic editing goal is object removal. One can remove objects in such a way that the answer remains invariant~(wherein only objects irrelevant to the QA are removed) as shown in Figure \ref{fig:teaser}~(top/middle). Alternately one could also make covariant edits where we remove the object mentioned in the QA and hence expect the answer to change in a predictable manner as shown in Figure \ref{fig:teaser}~(bottom). We explore both invariant and covariant forms of editing and quantify how consistent models are under these edits.

We employ a GAN-based \cite{rshetty_gan} re-synthesis model to automatically remove objects.
Our data generation technique helps us create exact complementary pairs of the image as shown in Figures \ref{fig:teaser}, \ref{fig:pos_neg }. We pick three recent models which represent different approaches to VQA to analyze robustness: a simple CNN+LSTM (CL) model, an attention-based model~(SAAA \cite{SAAA}) and a compositional model (SNMN \cite{SNMN}). We show that all the three models are brittle to semantic variations in the image, revealing the false correlation that the models exploit to predict the answer. Furthermore, we show that training data augmentation with our synthetic set can improve models robustness.

Our motivation to create this complementary dataset stems from the desire to study how accurate and consistent different VQA models are and to improve the models by the generated `complementary' data (otherwise not available in the dataset). While data  augmentation and cyclic consistency are making the VQA models more robust \cite{kafle_data_aug, data_aug_sunny_dark_outside, marcus_paper} towards the natural language part, we take a step forward to make the models consistent to semantic variations in the images. We summarize our main contributions as follows:

\begin{itemize}[leftmargin=*,itemsep=0.3pt,topsep=1pt]
	\item We propose a novel approach to analyze and quantify issues of VQA models due to spurious correlation and biases of data and models. We use synthetic data to quantify these problems with a new metric that measures erroneous inconsistent predictions of the model.
    \item We contribute methodology and a synthetic dataset~\footnote{https://rakshithshetty.github.io/CausalVQA/} that complements VQA datasets by systematic variations that are generated by our semantic manipulations. We complement this dataset by a human study that validates our approach and provides additional human annotations. %
	\item We show how the above-mentioned issues can be reduced by a data augmentation strategy - similar to adversarial training. We present consistent results across a range of questions and three state of the art VQA methods and show improvements on synthetic as well as real data.
	\item  While we investigate diverse question types, we pay particular attention to counting by creating an covariant edited set and show that our data augmentation technique can also improve counting robustness in this setting.
\end{itemize}

\section{Related Work}
\myparagraph{Visual Question Answering} 
There has been growing interest in VQA~\cite{survey_1, survey_2} recently, which can be attributed to the availability of large-scale datasets \cite{vqa_v2, GQA, vqa_v1, DAQUAR, imagenet_cvpr09} and deep learning driven advances in both vision and NLP.
There has been immense progress in building VQA models \cite{deep_lstm_cnn, AskYourNeurons, vqa_joint_embed_1, vqa_joint_embed_2} using LSTMs \cite{LSTM} and convolutional networks \cite{CNN_LeCun, ResNet} to models that span different paradigms such as attention networks \cite{hie_coattn, SAAA, SAN} and compositional module networks \cite{NMN, N2NMN, SNMN, MAC}. In our work, we pick a representative model from each of these three design philosophies and study their robustness to semantic visual variations.

\myparagraph{Robustness in VQA} 
Existing VQA models often exploit language and contextual priors to predict the answers~\cite{yinyang,  vqa_existing_bias, vqa_v2, dont_just_assume}. To understand how much do these models actually see and understand, various works have been proposed to study the robustness of models under different variations in the input modalities. \cite{dont_just_assume} shows that changing the prior distributions for the answers across training and test sets significantly degrades models' performance. \cite{data_aug_sunny_dark_outside, marcus_paper} study the robustness of the VQA models towards linguistic variations in the questions. They show how different re-phrasings of the questions can cause the model to switch their answer predictions. In contrast, we study the robustness of VQA models to semantic manipulations in the image and propose a data augmentation technique to make the models robust.

\myparagraph{Data Augmentation for VQA} Data Augmentation has been used in VQA to improve model's performance either in the context of accuracy \cite{kafle_data_aug} or robustness against linguistic variations in questions \cite{data_aug_sunny_dark_outside, marcus_paper}. \cite{kafle_data_aug} generated new questions by using existing semantic annotations and a generative approach via recurrent neural network. They showed that augmenting these questions gave a boost of around 1.5\% points in accuracy.  \cite{marcus_paper} propose a cyclic-consistent training scheme where they generate different rephrasings of question (based on answer predicted by the model) and train the model such that answer predictions across the generated and the original question remain consistent. \cite{data_aug_sunny_dark_outside}  proposes a data augmentation module that automatically generates entailed (or similar-intent) questions for a source QA pair and fine-tunes the VQA model if the VQA's answer to the entailed question is consistent with the source QA pair.

\section{Synthetic Dataset for Variances and Invariances in VQA}
While robustness w.r.t linguistic variations \cite{marcus_paper, data_aug_sunny_dark_outside} and changes in answer distributions \cite{dont_just_assume} have been studied, we explore how robust VQA models are to semantic changes in the images. 
For this, we create a synthetic dataset by removing objects irrelevant and relevant to the QA pairs and propose consistency metrics to study the robustness. Our dataset is built upon existing VQAv2 \cite{vqa_v2} and MS-COCO \cite{ms_coco_dataset} datasets. We target the 80 object categories present in the COCO dataset \cite{ms_coco_dataset} and utilize a GAN-based \cite{rshetty_gan} re-synthesis technique to remove them. The first key step in creating this dataset is to select a candidate object for removal for each Image-Question-Answer~(IQA) pair. Next, since we use an in-painter-based GAN, we need to ensure the removal of the object does not affect the quality of the image or QA in any manner. We introduce vocabulary mapping to take care of the former and area-overlapping criteria for the latter. We discuss these steps in detail to generate the edited set in irrelevant removal setting and later extend these to relevant object removal.

\subsection{InVariant VQA (\vqaiv{}) }%
For the creation of this dataset, we select and remove the objects irrelevant to answering the question. Hence the model is expected to make the same predictions on the edited image. A change in the prediction would expose the spurious correlations that the model is relying on to answer the question. Some examples of the semantically edited images along with the original images can be seen in Figures \ref{fig:teaser}, \ref{fig:pos_neg }. For instance, in Figure \ref{fig:pos_neg }~(top-right), for the question about the color of the surfboard, removing the person should not influence the model's prediction. In order to generate the edited image, we first need to identify person as a potential candidate which in turn requires studying the objects present in the image and the ones mentioned in the QA. Since we use VQA v2 dataset \cite{vqa_v2}, where all the images overlap with MS-COCO \cite{ms_coco_dataset}, we can access the ground-truth bounding box and segmentation annotations for each image. In total, there are 80 different object classes in MS-COCO which become our target categories for removal.

\myparagraph{Vocabulary mapping} To decide if we can remove an object, we need to first map all the object referrals in question and answer onto the 80 COCO categories. These categories are often addressed in the QA space by many synonyms or a subset representative of that class. For example- people, person, woman, man, child, he, she, biker all refer to the category: `person'; bike, cycle are commonly used for the class `bicycle'. To avoid erroneous removals, we create an extensive list mapping nouns/pronouns/synonyms used in the QA vocabulary to the 80 COCO categories. Table \ref{table:vocab_matching} shows a part of the object mapping list.
The full list can be found in supplementary material, section A.1.

 \begin{table}
\small
\centering
\begin{tabular}{ l  l } 
 \toprule
 COCO categories & Additional words  mapped \\ [0.5ex]
 \midrule
 person & man, woman, player, child, girl, boy \\
  & people, lady, guy, kid, he etc \\
 fire hydrant & hydrant, hydrate, hydra \\ 
 wine glass & wine, glass, bottle, beverage, drink \\
 donut & doughnut, dough, eating, food, fruit \\
 chair & furniture, seat \\ [0.2ex]
 ... & ...\\
 \bottomrule
\end{tabular}
\caption{Example of vocabulary mapping from QA space to COCO categories. If any of these words (in the right column) occur in the QA, these words are mapped to the corresponding COCO category (in the left column).}
\vspace{-4mm}
\label{table:vocab_matching}
\end{table}

Let $O_{\text{\it I}}$ represent the objects in the images (known via COCO segmentations), $O_{\text{\it QA}}$ represent the objects in the question-answer (known after vocabulary mapping). Then our target object for removal, $O_{\text{\it target}}$, is given by $O_{\text{\it I}} - \{O_{\text{\it I}} \cap O_{\text{\it QA}}\}$. We assume that if the object is not mentioned in the QA, it is not relevant and hence can be safely removed.

\myparagraph{Area-Overlap threshold} The next step is to make sure that the removal of $O_{\text{\it target}}$ does not degrade the quality of the image or affect the other objects mentioned in the QA. Since we use an in-painter based GAN \cite{rshetty_gan}, we find that larger object removal is harder to in-paint leaving the images heavily distorted. In order to avoid such distorted images, we only remove the object if the area occupied by its largest instance is less than 10\% of the image area. Furthermore, we also consider if the object being removed overlaps in any manner with the object that is mentioned in the QA. We quantitatively measure the overlap score as shown in Equation \ref{eq:overlap_score} where $M_{\text{\it O}}$ denotes the dilated ground truth segmentation mask of all the instances of the object. We only remove the object if the overlap score is less than 10\%.

\begin{equation}
\setlength\abovedisplayskip{0pt}
     \text{Overlap score} (O_{\mathrm{target}}, O_{\mathrm{QA}}) = \frac{(M_O)^{\mathrm{target}}\cap (M_{O})^{\mathrm{QA}}}{(M_{O})^{\mathrm{QA}}}
     \label{eq:overlap_score}
\end{equation}

\mario{no text in math mode. either $O_{\text{target}}$ or $O_{\text{\it target}}$ }

\myparagraph{Uniform Ground-Truth} Finally, we only aim to target those IQAs which have uniform ground-truth answers. In VQA v2 \cite{vqa_v2}, all the questions have 10 answers, while it is good to capture diversity in open-ended question-answering, it also introduces ambiguity, especially in case of counting and binary question types.  To avoid this ambiguity in our robustness evaluation, we build our edited set by only selecting to semantically manipulate those IQs which have a uniform ground truth answer.  

Finally, we remove all the instances of the target object from the image for those IQAs which satisfy the above criteria using the inpainter GAN \cite{rshetty_gan}. We call our edited set as \vqaiv{} as removal of objects does not lead to any change in answer, the answer is invariant to the semantic editing. Table \ref{table:edited_set_stats_ALL_ANS_SAME} shows the number of edited IQAs in \vqaiv{}.  
While our algorithm involves both manually curated heuristics to select the objects to remove, and a learned in-painter-based GAN model to perform the removal, the whole pipeline is fully automatic. 
This allows us to apply it to the large-scale VQA dataset with 658k IQA triplets.

\begin{table}
\centering
\small
\begin{tabular}{l  l l  l  l l l} 
 \toprule
 & \multicolumn{3}{c}{ \vqaiv{} }  & \multicolumn{3}{c}{ \vqacv{}}\\ [0.5ex]
 \midrule
 \#IQA  &  train   & val & test  & train & val & test \\ [0.5ex]
 \midrule
 
 real &   148013 & 7009 &63219  & 18437 & 911 &8042 \\ 
 realNE & 42043 &2152 &18143 & 13035 &648  &5664\\ 
 edit & 256604  &11668 &108239 & 8555 &398 &3743\\ [0.2ex]
 \bottomrule
\end{tabular}
\caption{ \vqaiv{} and \vqacv{} distribution. Real refers to VQA \cite{vqa_v2} IQAs with uniform answers, realNE refers to IQAs for which no edits are possible (after vocabulary mapping and area-overlap threshold), edit refers to the edited IQA. We split the VQA val into 90:10 ratio, where the former is used for testing purpose and latter for validation.}
\vspace{-3mm}
\label{table:edited_set_stats_ALL_ANS_SAME}
\end{table}

\myparagraph{Validation by Humans}
We get a subset (4.96k IQAs) of our dataset validated by three humans.
The subset is selected based on an inconsistency analysis of 3 models covered in the next Section \ref{robustness_flip_desc}.
Every annotator is shown the edited IQA and is asked to say if the answer shown is correct for the given image and question (yes/no/ambiguous).
According to the study, 91\% of the time all the users agree that our edited IQA holds.
More details about the study are in the supplementary material (section A.2).
\subsection{CoVariant VQA (\vqacv{})} %

An alternate way of editing images is to target the object in the question. Object-specific questions like counting, color, whether the object is present or not in the image are suitable for this type of editing. We choose counting questions where we generate complementary images with one instance of the object removed. If the models can count $n$ instances of an object in the original image, it should also be able to count $n-1$ instances of the same object in the edited image. Next, we will describe how to generate this covariant data for counting.

First, we collect all the counting questions in the VQA set: selecting questions which contained words `many' and `number of' and which had numeric answers. Next, we focus on removing instances of the object which is to be counted in the question. 
Vocabulary mapping is used to identify the object mentioned in the question $O_{Q}$. Then only those images are retained where the number of the target object instances according to COCO segmentations match the IQA ground-truth answer $A$ given by 10 human annotators.

For the generation of this set, we use the area threshold as 0.1,  we only intend to remove the instance if it occupies less than 10\% of the image. Furthermore for overlap, since we do not want the removed instance to interfere with the other instances of the object, two masks considered to measure the score are: (1). dilated mask of instance to be removed (2). dilated mask of all the other instances of the object. The object is only removed if the overlap score is zero.

We call our edited set as \vqacv{} since removal of the object leads to a covariant change in answer. Table \ref{table:edited_set_stats_ALL_ANS_SAME} shows the number of edited IQAs in VQA-CV. Figure \ref{fig:pos_neg }~(bottom row) shows a few examples from our edited set. We only target one instance at a time. More such visual examples can be found in supplementary (section B.2) . 
\section{Experiments: Consistency analysis}

\begin{figure*}
\centering
\small
  \begin{tabular}{l c c  l c c}
    \Xhline{2\arrayrulewidth}
    \multicolumn{3}{c}{pos$\rightarrow$neg} &     \multicolumn{3}{c}{neg$\rightarrow$pos} \\
    \Xhline{2\arrayrulewidth}
    
    \multicolumn{3}{c}{\myit{Q}: What are the shelves made of?}   &  \multicolumn{3}{c}{\myit{Q}: What color is the surfboard?} \\
    
    \multicolumn{2}{c}{\myit{A}: glass} & \multicolumn{1}{c}{\textit{vases removed}; \myit{A}: glass}
    
    &     \multicolumn{2}{c}{\myit{A}: white} & \multicolumn{1}{c}{\textit{person removed}; \myit{A}: white}\\
    
    \multicolumn{3}{c}{\includegraphics[height=0.2\textwidth,  width=0.4\textwidth, trim= 0 0 0 0, clip]{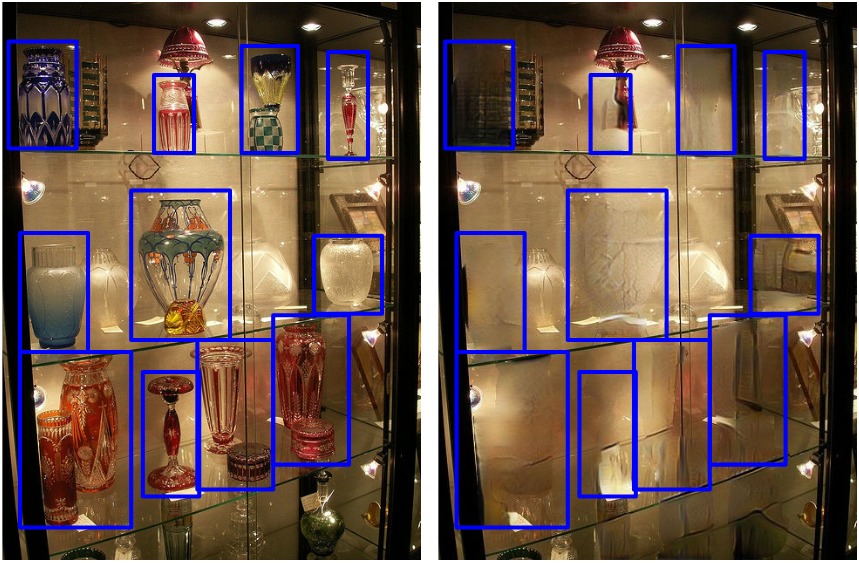}} &     \multicolumn{3}{c}{\includegraphics[height=0.2\textwidth, width=0.48\textwidth, trim= 0 0 0 0, clip]{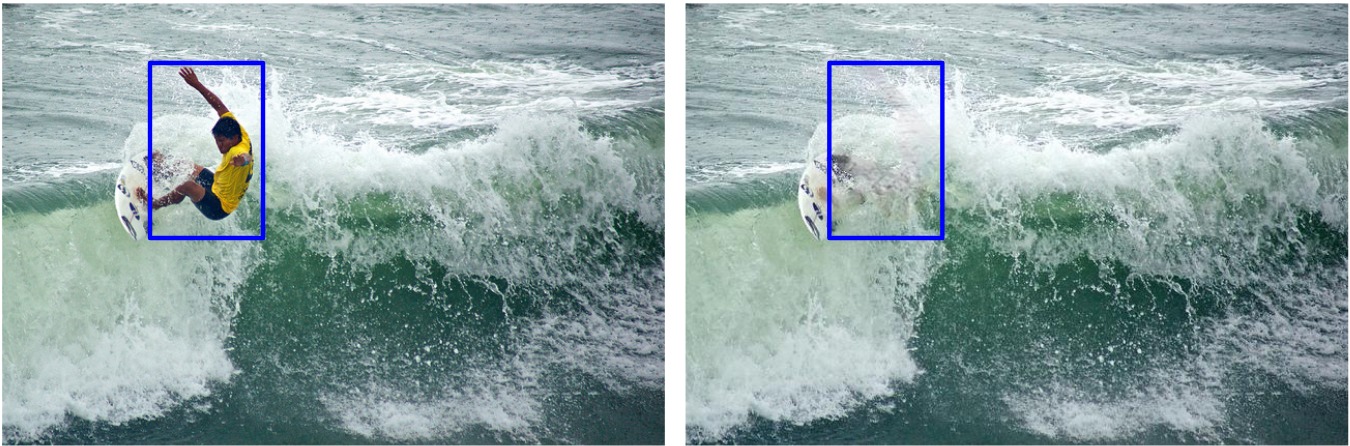}}\\
    CNN+LSTM & \green{glass} & \red{wood} &          CNN+LSTM & \red{yellow}  & \green{white}  \\
    SAAA & \green{glass} &  \red{metal}  &     SAAA  & \green{white}   & \green{white}   \\ 
    SNMN & \green{glass} & \red{metal}  &     SNMN &  \red{yellow}  & \green{white}   \\
    \Xhline{2\arrayrulewidth}
   
    \multicolumn{3}{c}{ \myit{Q}: Are there zebras in the picture? }  & \multicolumn{3}{c}{\myit{Q}: Is there a cat?} \\
    
    \multicolumn{2}{c}{\myit{A}: yes} & \multicolumn{1}{c}{\tab \textit{giraffes removed}; \myit{A}: yes}
    
    &    \multicolumn{2}{c}{\myit{A}: no} & \multicolumn{1}{c}{\tab \textit{dogs removed}; \myit{A}: no}\\
        
    \multicolumn{3}{c}{\includegraphics[width=0.48\textwidth, trim= 0 0 0 0, clip]{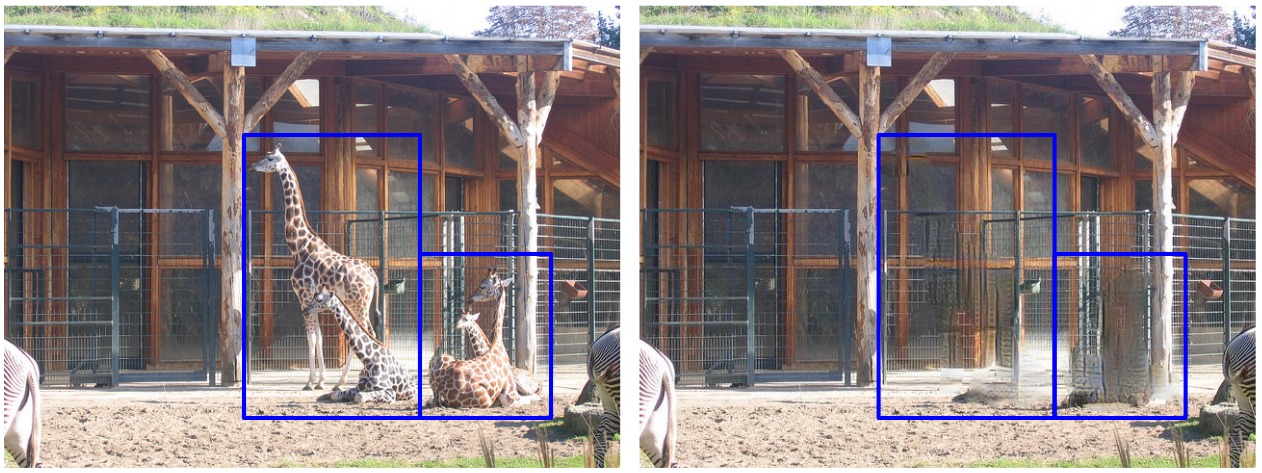}}
    &\multicolumn{3}{c}{\includegraphics[width=0.5\textwidth, trim= 0 0 0 0, clip]{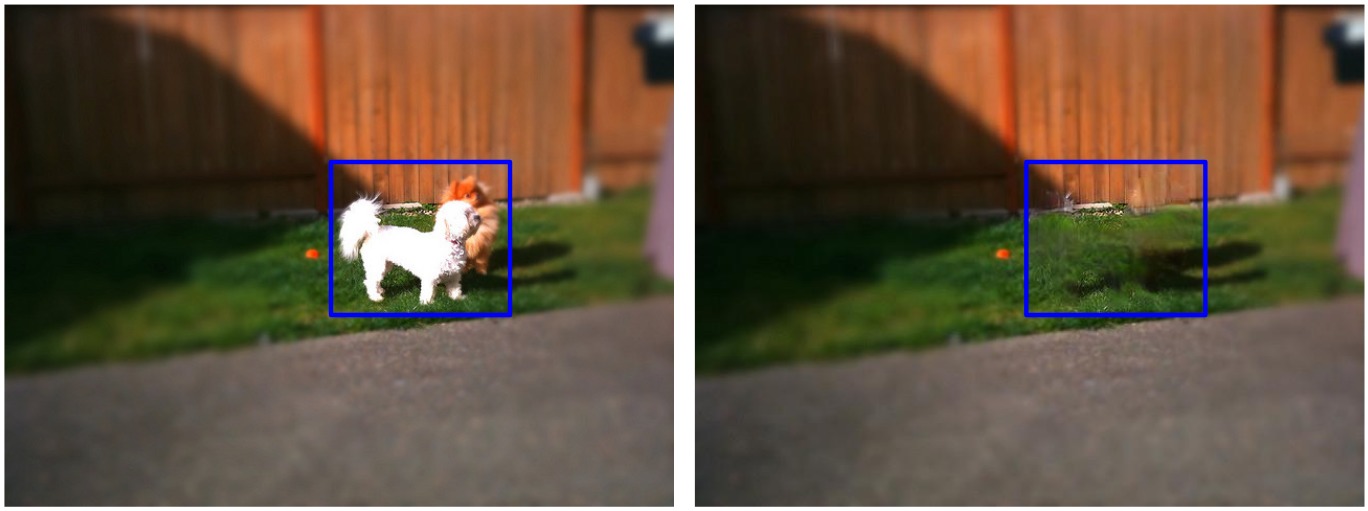}}\\
        
    CNN+LSTM & \green{yes} & \red{no} &     CNN+LSTM & \red{yes}  & \green{no}  \\
    SAAA & \green{yes } & \red{no} &    SAAA & \red{yes}  & \green{no}  \\
    SNMN & \green{yes} & \red{no}  &    SNMN & \red{yes}  & \green{no}  \\
    
    \Xhline{2\arrayrulewidth} 
    \multicolumn{3}{c}{\myit{Q}: What sport is he playing? } 
    &     \multicolumn{3}{c}{\myit{Q}: What room of a house is this?} \\
    
    \multicolumn{2}{c}{\myit{A}: soccer} & \multicolumn{1}{c}{\tab[1cm] \textit{sports-ball}; \myit{A}: soccer}
    
    &     \multicolumn{2}{c}{\myit{A}: kitchen} & \multicolumn{1}{c}{\textit{bowl}; \myit{A}: kitchen}\\

    \multicolumn{3}{c}{\includegraphics[width=0.5\textwidth, trim= 0 0 0 0, clip]{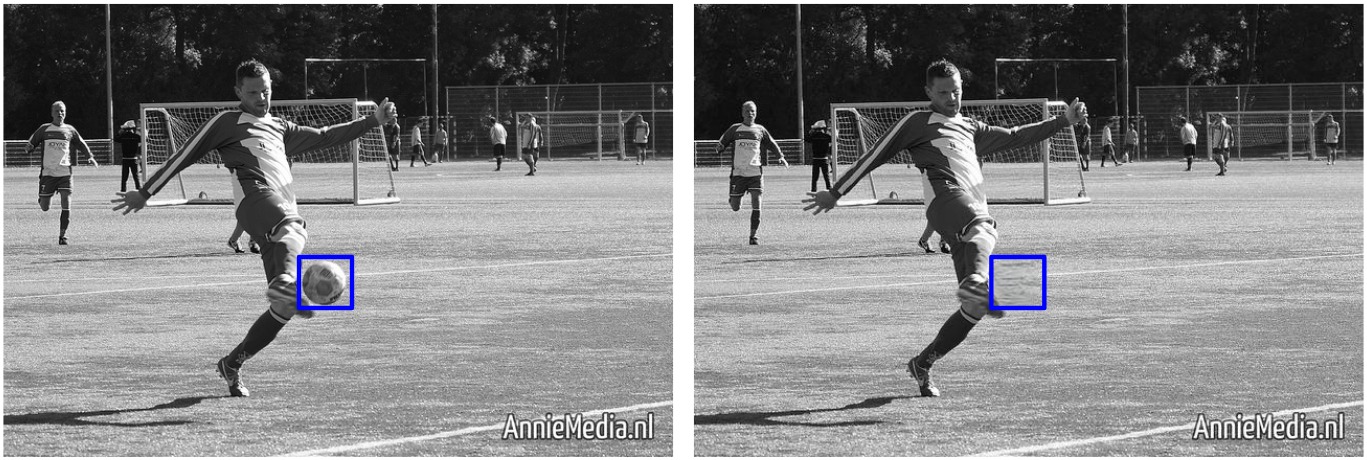}} 
     &     \multicolumn{3}{c}{\includegraphics[width=0.48\textwidth, trim= 0 20 0 0,
    clip]{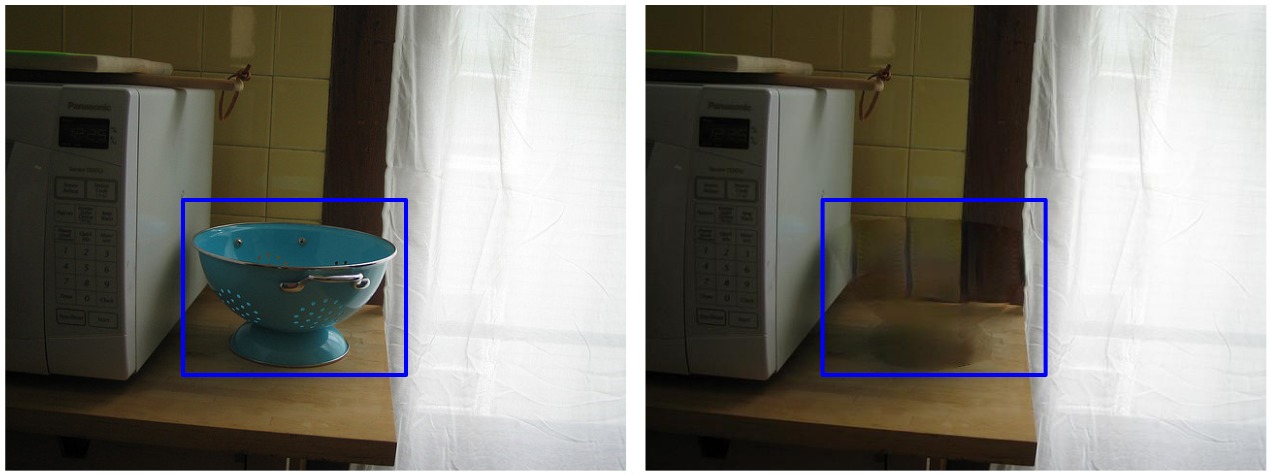}}\\
    
    CNN+LSTM & \green{soccer} & \textcolor{red}{tennis}   &
    CNN+LSTM & \red{bathroom} & \green{kitchen}  \\
    SAAA & \green{soccer} & \textcolor{red}{tennis}  &     SAAA  & \red{bathroom} & \green{kitchen}   \\
    SNMN & \green{soccer} & \textcolor{red}{tennis} &     SNMN &  \red{bathroom} & \green{kitchen}  \\

    \Xhline{2\arrayrulewidth}   
    \multicolumn{3}{c}{\myit{Q}: How many dogs are there? } 
    &  \multicolumn{3}{c}{\myit{Q}: How many giraffe are there?}  \\
    
    \multicolumn{2}{c}{\myit{A}: 1}  &    
    \multicolumn{1}{c}{\tab \textit{dog removed}; \myit{A}: 0}  & 
    
    \multicolumn{2}{c}{\myit{A}:3}  &    
    \multicolumn{1}{c}{\tab \textit{ giraffe removed}; \myit{A}: 2}  \\
    
    \multicolumn{3}{c}{\includegraphics[width=0.48\textwidth, trim= 0 0 0 0, clip]{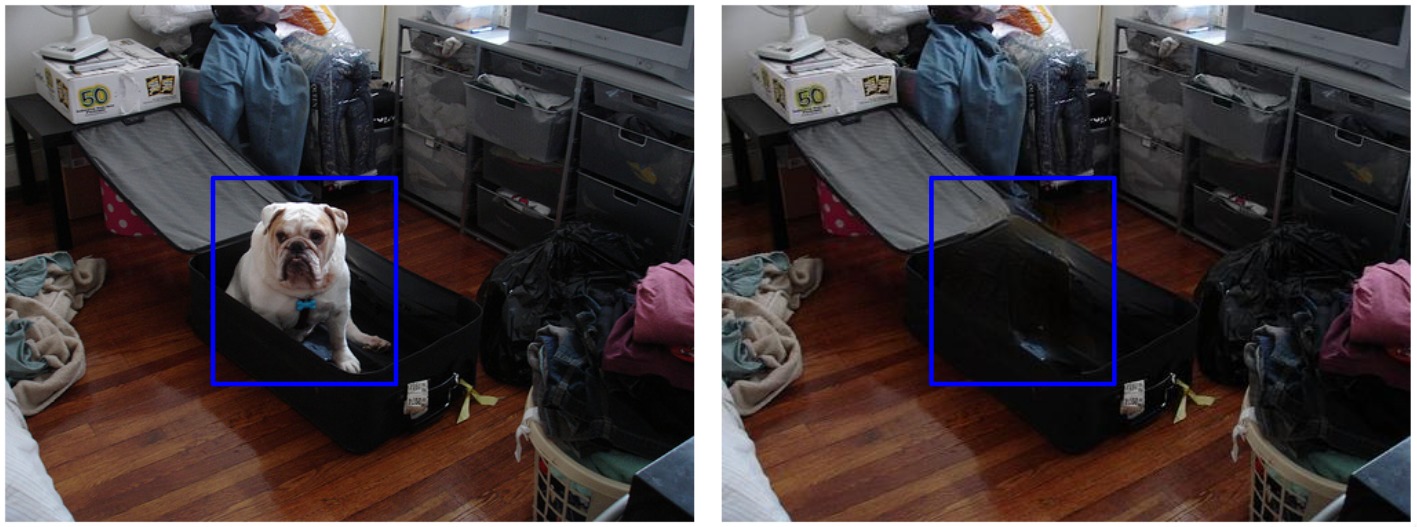}} & 
    \multicolumn{3}{c}{\includegraphics[width=0.48\textwidth, trim= 0 0 0 0, clip]{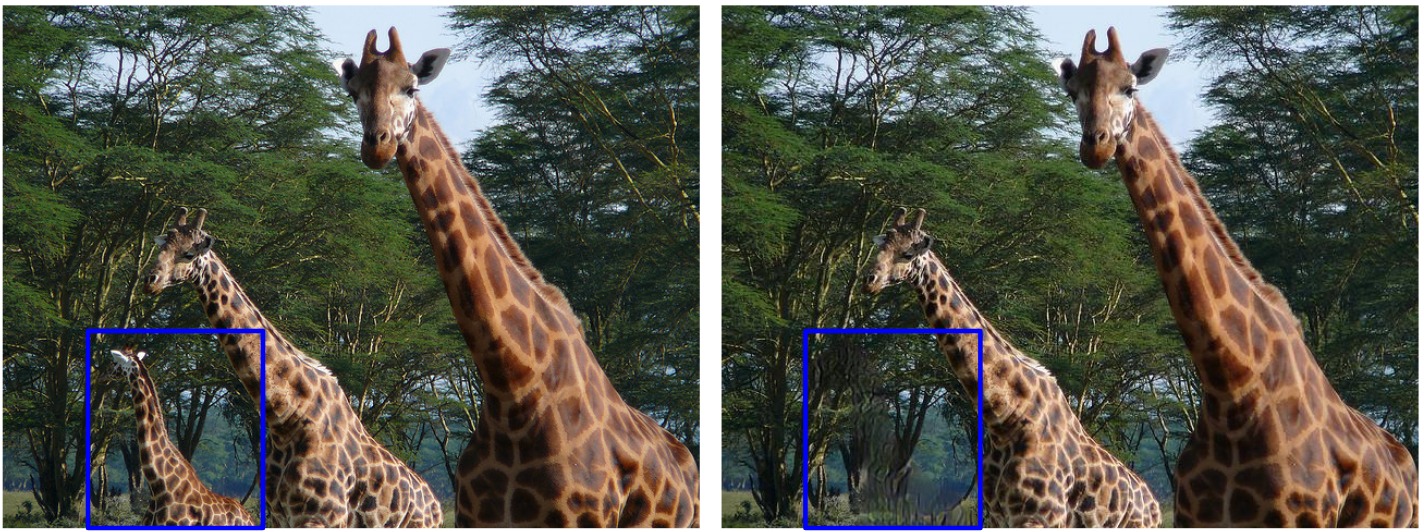}}\\
    
    CNN+LSTM & \green{1}  & \red{2}    & CNN+LSTM & \red{1} & \green{2} \\
    SAAA  & \green{1} & \red{1}    &     SAAA & \red{2} &  \green{2}  \\
    SNMN &  \green{1} & \red{1} &   SNMN & \red{2} & \green{2}  \\
    \Xhline{2\arrayrulewidth}   
  \end{tabular}
  \caption{Existing VQA models exploit spurious correlations to predict the answer often looking at irrelevant objects. Shown above are the predictions for 3 different VQA models on  original and edited images from our synthetic datasets \vqaiv{} and \vqacv{}.  \mario{I think the examples look great.}}
  \label{fig:pos_neg }
\end{figure*}

The goal of creating edited datasets is to gauge how consistent are the models to semantic variations in the images.
In \vqaiv{}, where we remove objects irrelevant to the QA from the image, we expect the models predictions to remain unchanged. 
In \vqacv{}, where one of the instances to be counted is removed, we expect the predicted answer to reduce by one as well.  
Next, we briefly cover the models' training and then study their performances both in terms of accuracy and consistency. 
We propose consistency metrics based on how often the models flip their answers and study the different type of flips qualitatively and quantitatively.

\vspace{2mm}

\myparagraph{VQA models and training}
For comparison and analysis, we select three models from the literature, each representing a different design paradigm: a simple CNN+LSTM (CL)  model, an attention-based model (SAAA \cite{SAAA}) and a compositional model (SNMN \cite{SNMN}).
We use the official code for training the SNMN \cite{SNMN} model, \cite{github_snmn}. 
SAAA \cite{SAAA} is trained using the code available online \cite{pytorch_vqa_SAAA}. 
We modified this SAAA code in order to get CL model by removing the attention layers from the network. 
As we use the VQA v2 val split for consistency evaluation and testing, the models are trained using only the train split.
Further details of these models and hyper-parameters used can be found in the supplementary (section B.1). Table \ref{table:accuracy_val_vqa} shows the accuracy scores on VQA v2 val set for models trained by us along with similar design philosophy models benchmarked in \cite{dont_just_assume} and \cite{vqa_v2}. The models chosen by us exceed the performance of other models within the respective categories.

\begin{table}[t]
\small
\centering
\begin{tabular}{l l  l l   } 
 \toprule
 \multicolumn{2}{c}{Trained by us} & \multicolumn{2}{c}{For comparison}\\
 \midrule
 CNN+LSTM &53.32 & d-LSTM Q + norm I  \cite{deep_lstm_cnn} & 51.61 \\
 \midrule
 SAAA \cite{SAAA} &61.14 & SAN \cite{SAN} & 52.02 \\
 & &  HieCoAttn \cite{hie_coattn} & 54.57 \\
 & & MCB \cite{MCB} & 59.71 \\
 \midrule
 SNMN \cite{SNMN} & 58.34 & NMN \cite{NMN} & 51.62  \\ [0.2ex]
 \bottomrule
\end{tabular}
\caption{ Accuracy (in \%) of different models when trained on VQA v2 train and tested on VQA v2 val.}
\vspace*{-4mm}
\label{table:accuracy_val_vqa}
\end{table}

\label{robustness_flip_desc}

\myparagraph{Consistency} The edited data is created to study the robustness of the models. 
Since we modify the images in controlled manner, we expect the models predictions to stay consistent. 
Robustness is quantified by measuring how often models change their predictions on the edited IQA from the prediction on original IQ.
On \vqaiv{}, a predicted label is considered ``flipped'' if it differs from the prediction on the corresponding unedited image. 
On \vqacv{}, if the answer on the edited samples is not one less than the prediction on original image, it is considered to be ``flipped''.  %

We group the observed inconsistent behavior on edited data into three categories: 1. neg$\rightarrow$pos 2. pos$\rightarrow$neg 3. neg$\rightarrow$neg. neg$\rightarrow$pos flip means that answer predicted on the edit IQA was correct but the prediction on the corresponding real IQA was wrong. Other flips are defined analogously. In the neg$\rightarrow$neg flip, answer predicted is wrong in both the cases. 
While all forms of label flipping show inconsistent behaviour, the pos$\rightarrow$neg and neg$\rightarrow$pos categories are particularly interesting. 
In these the answer predicted is correct before and afterward the edit, respectively. 
These metrics show that there is brittleness even while making correct predictions and indicate that models exploit spurious correlations while making their predictions.

\myparagraph{Quantitative analysis}
Table \ref{table:flipping_0.1_0.1} shows the accuracy along with the consistency numbers for all the 3 models on the \vqaiv{} test split. Consistency is measured across edited \vqaiv{} IQAs and corresponding real IQAs from VQA v2. Accuracy is reported on real data from VQA v2 (original IQAs with uniform answers). We follow this convention throughout our paper.
On the original data, we see that SAAA is the most accurate model (70.3\%) as compared to SNMN (66\%) and CL (60.2\%).  
In terms of robustness towards the variations in the images, CL model is the least consistent- with a 17.9\% flipping on the edit set compared to the predictions on the corresponding original IQA.
For SAAA, 7.85\% flips, making SNMN the most robust model with 6.522\% flips.
SAAA and SNMN are much more stable than CL. 
A point noteworthy here is that SNMN turns out to be the most robust despite its accuracy being lesser than SAAA. 
This shows that higher accuracy does not necessarily mean we have the best model, further highlighting the need to study and improve the robustness of the models.
Of particular interest are the pos$\rightarrow$neg and neg$\rightarrow$pos scores, which are close to 7\% each for the CL model. For a neg$\rightarrow$pos flip, the answer to change from an incorrect answer to one correct answer of the 3000 possible answers (size of answer vector). If the removed object was not used by the model, as it should be, and editing caused uniform perturbations to the model prediction, this event would be extremely rare ($p(\text{neg}\rightarrow\text{pos}) = 1/3000 * 39.8 = 0.013\%$). However we see that this occurs much more frequently~(6.9\%), indicating that in these cases model was spuriously basing its predictions on the removed object and thus changed the answer when this object was removed.

In the \vqacv{} setting, where we target counting and remove one instance of the object to be counted, we expect the models to maintain n/n-1 consistency on real/edited IQA. 
As we see from Table \ref{table:flipping_del}, the accuracy on orig set is quite low for all the models reflecting the fact that counting is a hard problem for VQA models. 
SAAA (49.9\%) is the most accurate model with SNMN at 47.9\% and CL at 39.4\%. 
In terms of robustness, we see that for all 3 models are inconsistent more than 75\%, meaning for >75\% for the edited IQAs, if models could correctly count n objects in the original IQA, it wasn't able to count n-1 instances of the same object in the edited IQA.
These numbers further reflect that counting is a difficult task for VQA models and enforcing consistency on it seems to break all 3 models.
\begin{table}
\small
\centering
\begin{tabular}{l  c c c  } 
\toprule 
 & CL (\%) & SAAA (\%) & SNMN (\%) \\ 
 \midrule
 Accuracy orig & 60.21 & 70.26 & 66.04 \\
 \midrule
 Predictions flipped & 17.89 & 7.85 & 6.52\\
 pos$\rightarrow$neg & 7.44 & 3.47 & 2.85  \\
 neg$\rightarrow$pos & 6.93 & 2.79 & 2.55 \\
 neg$\rightarrow$neg & 3.53 & 1.58 & 1.12 \\ [0.2ex]
 \bottomrule
\end{tabular}
\caption{ Accuracy-flipping on real data/\vqaiv{} test set.} %
\vspace*{-3mm}
\label{table:flipping_0.1_0.1}
\end{table}
\begin{table}
\small
\centering
\begin{tabular}{l  c c c  } 
\toprule
 & CL (\%) & SAAA (\%) & SNMN (\%) \\ 
 \midrule
 Accuracy orig & 39.38 & 49.9 & 47.948 \\

 \midrule
 Predictions flipped & 81.41 & 78.44 & 78.92 \\
 pos$\rightarrow$neg & 28.69  &31.66 & 32.35\\
 neg$\rightarrow$pos & 20.57 & 25.38 & 23.51\\
 neg$\rightarrow$neg & 32.14  &21.4 & 23.06 \\ [0.2ex]
 \bottomrule
\end{tabular}
\caption{ Accuracy-flipping on real data/\vqacv{} test set.} %
\vspace*{-3mm}
\label{table:flipping_del}
\end{table}
In the next section, we discuss these flips with some visual examples.

\myparagraph{Qualitative analysis}
We visualize the predictions of the models on a few original and edited IQAs for all the 3 models in Figure \ref{fig:pos_neg }. The left half shows examples of pos$\rightarrow$neg and the right half shows the neg$\rightarrow$pos flips. Existing VQA models often exploit false correlations to predict the answer. We study the different kinds of flips in detail here and see how they help reveal these spurious correlations.

\myparagraph{pos$\rightarrow$neg} 
VQA models more often rely on the contextual information/ background cues/ linguistic priors to predict the answer rather than the actual object in the question. For instance, removal of the glass vases from the shelves in Figure \ref{fig:pos_neg } (Top-left) from the image causes all 3 models to flip their answers negatively, perhaps models were looking at the wrong object (glass vases) to predict the material of the shelves that also happened to be glass. In absence of giraffes, models cannot seem to spot the occluded zebras anymore- hinting that maybe they are confusing zebras with giraffes. Removing the sports-ball from the field  make all 3 models falsely change their predictions to tennis without considering the soccer field or the players. In the bottom-left, we also see that if models were spotting the one dog rightly in the original image, on it's edited counterpart( with no dog anymore )- it fails to answer 0. 
Semantic edits impact the models negatively here exposing the spurious correlations being used by the models to predict the correct answer on the original image. These examples also show that accuracy should not be the only sole criterion to evaluate performance. A quick look at the Table \ref{table:flipping_0.1_0.1}  show that for \vqaiv{}, pos$\rightarrow$neg flips comprise a major chunk (>40\%) of all the total flips. For \vqacv{} (refer Table \ref{table:flipping_del}) , these flips are 28-32\% absolute- again reinforcing the fact that VQA models are far from learning to count properly.

\myparagraph{neg$\rightarrow$pos}
Contrary to above, semantic editing here helps correct the predictions, meaning removal of the object causes the model to switch its wrong answer to one right answer by getting rid of the wrong correlations. For instance, removing the pink umbrella helps models predict correctly the color of the balloon Figure \ref{fig:teaser}~(middle) \RS{repeating image from teaser}. In Figure \ref{fig:pos_neg } (second-right), removing the dogs leave no animals behind and hence models now can correctly spot the absence of cat- hinting that they were previously confusing cats and dogs. In absence of the bowl, models can identify the room as kitchen- shows that too much importance is given to the bowl (which is falsely correlated to bathroom) and not to the objects in the background such as microwave. Towards the bottom-right, we see that removing a giraffe helps all the 3 models now- it's hard to say what is the exact reason for the behaviour but it indeed reflects upon the inconsistent behaviour of the models. 
From Table \ref{table:flipping_0.1_0.1} we see that these flips also comprise a significant number of the total flips (>35\%) for all the models. For \vqacv{} (refer Table \ref{table:flipping_del}), these numbers are in range 20-25\%, showing that counting is easier for these models when spurious correlations are removed.
\myparagraph{neg$\rightarrow$neg} 
These flips where answers change show the inconsistent behavior of models as well but since both the answers are wrong- they are harder to interpret. But in the end goal of building robust models, we expect consistent behavior even when making incorrect predictions.

All these flips show that existing VQA models are brittle to semantic variations in images. While VQA models are getting steadily better in terms of accuracy, we also want our models to be robust to visual variations. We want VQA models to not just be accurate but use the right cues to answer correctly. Accuracy combined with consistency can help us understand the shortcomings of the models.
\RS{Can be significantly shortened}

\mario{I like the detailed discussion. some reviewers might not be convinced in the beginning and going through the examples in detail might help - but we have to see how much space we have.}

\section{Robustification by Data Augmentation}

In the previous section, we see that VQA models are brittle to semantic manipulations. While these flips expose the inconsistent behaviour, they also show the underlying scope of improvement for VQA models and can be used to make the models more robust. In order to leverage the variances brought in by the synthetic data, we finetune all the models using real and real+synthetic data. Our analysis shows that using synthetic data significantly reduces inconsistency across a variety of question types.


For fine-tuning experiments, we use a strict subset of \vqaiv{} with an overlap score of zero. The performance of all the baseline models on this strict subset remains similar to Table \ref{table:flipping_0.1_0.1} (refer supplementary- section C.1). For SNMN, the model trained using a learning rate of $1e^{-3}$ is unstable while fine-tuning and hence we use a lower learning rate $2.5e^{-4}$  to train the model and further finetune this model.

\begin{figure}
\centering
\begin{tabular}{c }
    \includegraphics[width=0.5\linewidth, trim= 0 0 0 0, clip]{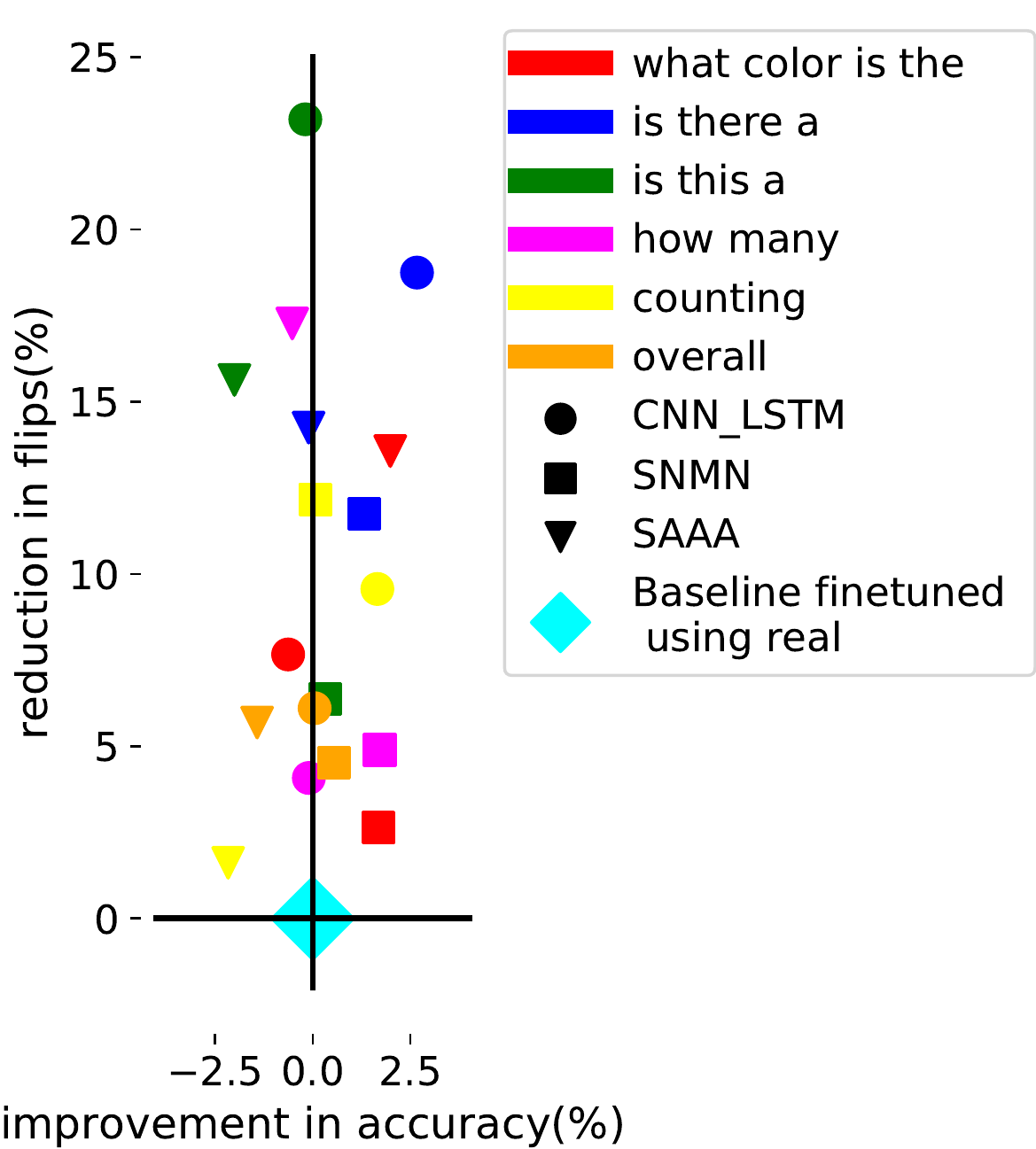}
    \includegraphics[width=0.5\linewidth, trim= 0 0 0 0, clip]{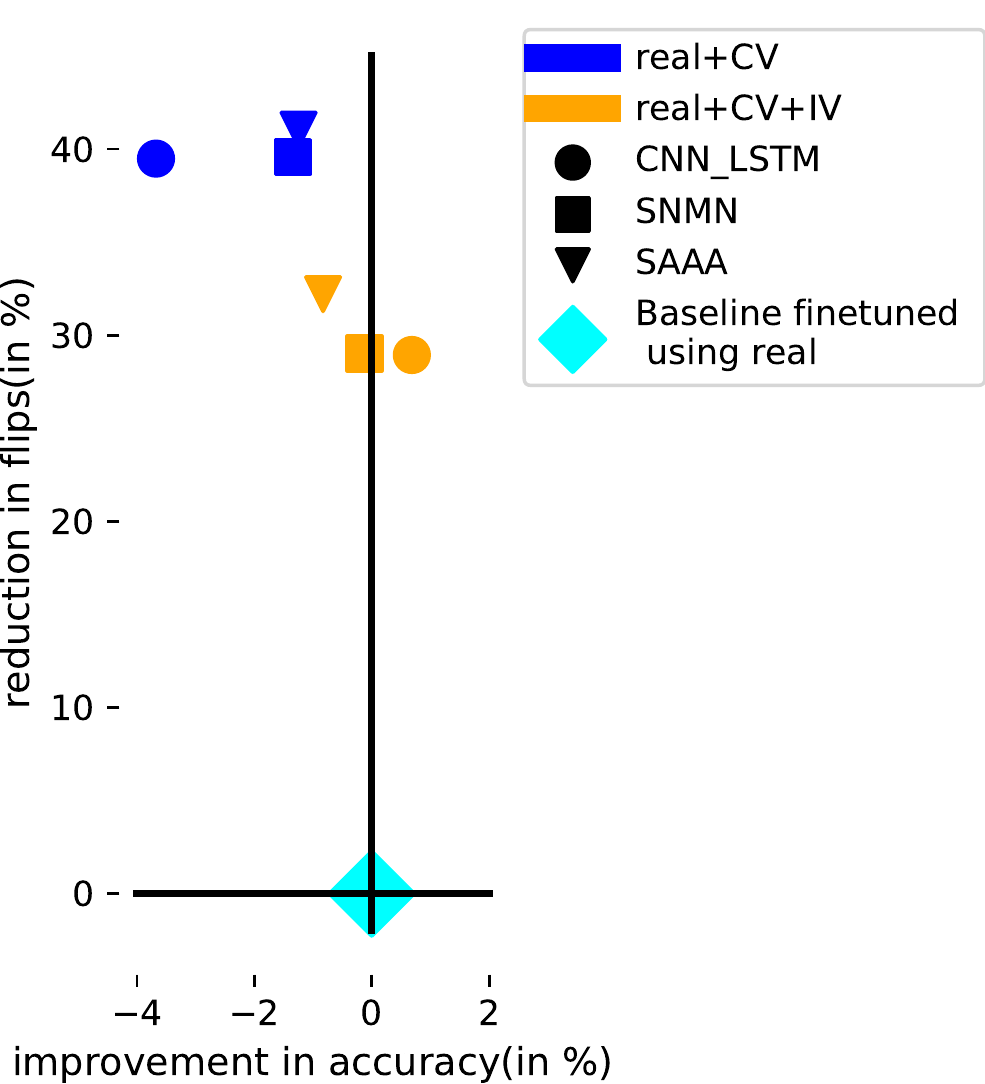}   \\
\end{tabular}
\caption{ Accuracy-flipping results of finetuning experiments. Plots show relative performance of models finetuned using real+edit data w.r.t to using just real data.}
\vspace*{-3mm}
\label{fig:DA_ques_types12}
\end{figure}





\begin{figure}
\centering
\small
  \begin{tabular}{l  c c c c}
    \Xhline{2\arrayrulewidth}   
    \multicolumn{5}{c}{\myit{Q}: What color is the mouse?} \\
    \multicolumn{3}{c}{ \myit{A}: white} & \multicolumn{2}{l}{ \textit{keyboards removed}; \myit{A}: white} \\
    \multicolumn{5}{c}{\includegraphics[width=0.43\textwidth, trim= 0 0 0 3cm, clip]{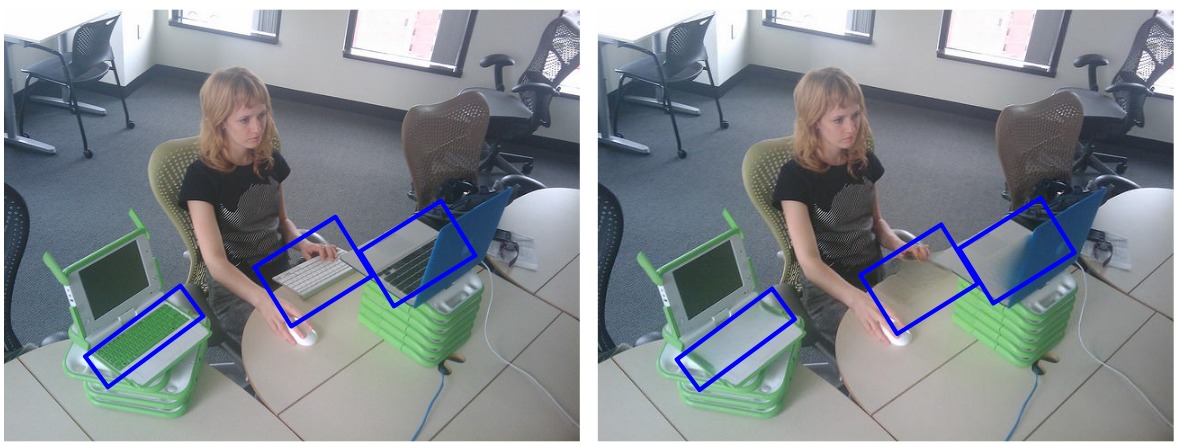}}\\
    & real & real+edit & \tab[0.7cm] real & real+edit\\ 
    CL & \green{white} & \green{white}  & \tab[0.7cm] \green{white} &\green{white} \\
    SAAA  & \red{green} & \green{white}  & \tab[0.7cm] \green{white} &\green{white}  \\ 
    SNMN &  \red{green} & \green{white}  & \tab[0.7cm] \green{white} &\green{white}  \\
    
    \Xhline{2\arrayrulewidth}   
    
    \multicolumn{5}{c}{\myit{Q}:Is there a bowl on the table?}\\
    \multicolumn{3}{c}{\myit{A}: no} & \multicolumn{2}{l}{\tab \textit{cup removed}; \myit{A}: no} \\
    \multicolumn{5}{c}{\includegraphics[width=0.43\textwidth, trim= 0 3cm 0 0cm, clip]{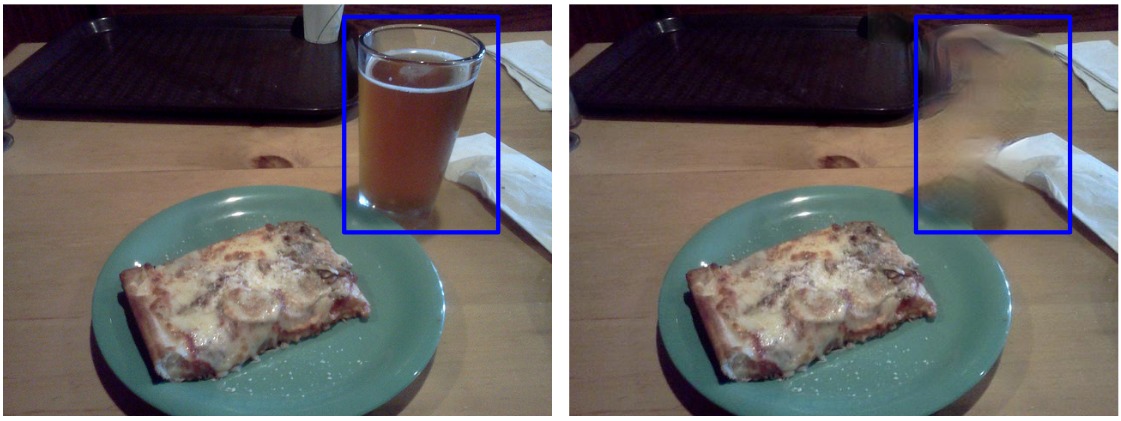}}\\
    & real & real+edit & \tab[0.7cm] real & real+edit\\ 
    CL & \green{no} & \green{no}  & \tab[0.7cm] \red{yes} &\green{no} \\
    SAAA  & \green{no} & \green{no}  & \tab[0.7cm] \red{yes} &\green{no}  \\ 
    SNMN &  \green{no} & \green{no}  & \tab[0.7cm] \red{yes} &\green{no}  \\
    
    \Xhline{2\arrayrulewidth}   

    \multicolumn{5}{c}{\myit{Q}: How many computer are there?} \\
    \multicolumn{3}{c}{\myit{A}: 2} &  \multicolumn{2}{l}{\tab \textit{dog removed}; \myit{A}: 2} \\
    \multicolumn{5}{c}{\includegraphics[width=0.43\textwidth, trim= 0 0 0 2cm, clip]{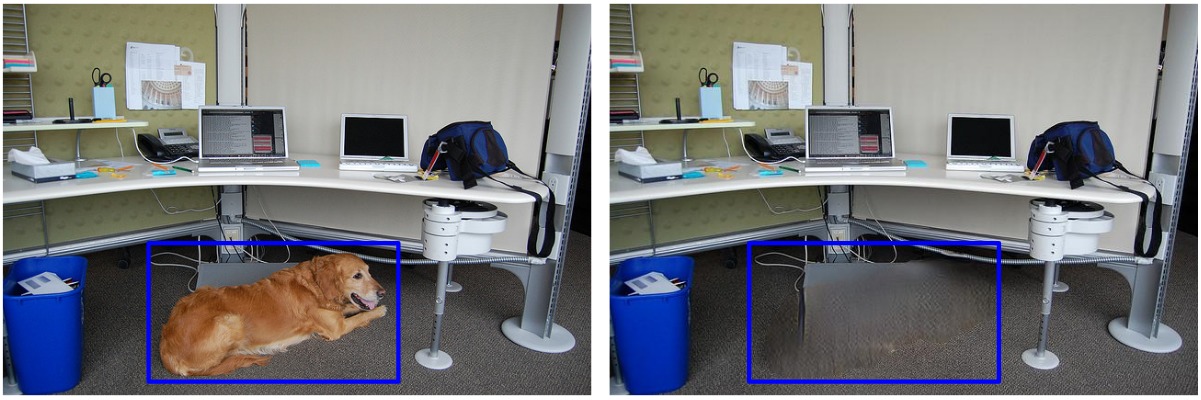}}\\
    & real & real+edit & \tab[0.7cm] real & real+edit\\ 
    CL & \green{2} & \green{2}  & \tab[0.7cm] \red{1} &\green{2} \\
    SAAA  & \red{1} & \green{2}  & \tab[0.7cm] \green{2} &\green{2}  \\ 
    SNMN &  \green{2} & \green{2}  & \tab[0.7cm] \red{1} &\green{2}  \\
    
    
    \Xhline{2\arrayrulewidth}   
    \multicolumn{5}{c}{Q: How many people are in the water?} \\
    \multicolumn{3}{c}{\myit{A}: 1} & \multicolumn{2}{l}{\tab \textit{person removed}; \myit{A}: 0} \\
    \multicolumn{5}{c}{\includegraphics[width=0.43\textwidth, trim= 0 0 0 1cm, clip]{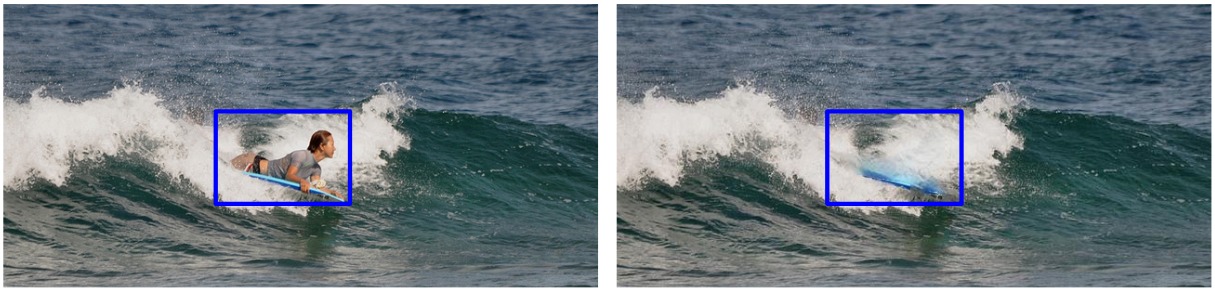}}\\
    & real & real+edit & \tab[0.7cm] real & real+edit\\ 
    CL & \green{1} & \green{1}  & \tab[0.7cm] \red{1} &\green{0} \\
    SAAA  & \green{1} & \green{1}  & \tab[0.7cm] \red{1} &\green{0}  \\ 
    SNMN &  \green{1} & \green{1}  & \tab[0.7cm] \red{1} &\green{0}  \\
    \Xhline{2\arrayrulewidth}   
  \end{tabular}
  \vspace*{-1mm}
  \caption{Visualizations from fine-tuning experiments using real/real+edit. Using real+edit makes models more consistent and in these examples- also accurate.}
  \vspace*{-4mm}
  \label{fig:DA_visual_aid}
\end{figure}

\myparagraph{InVariant VQA Augmentation} 
In order to train and test different models, we aim at specific question types and see if we are able to boost the model's performance on that question type. We select 4 question types based on how much they are affected from editing (i.e total number of flips/ total number of original IQA per question type) and if that question category has significant number of flipped labels in order to ensure we have enough edited IQAs for finetuning. Hence, we select the given 3 question categories and run our experiments on these splits: 1. `what color is the' 2. `is there a' 3. `is this a' 4. `how many'. Additionally we focus on all the counting questions. All these specialized splits have around 6.3k-12.5k IQAs in the real train split with 10.8k-15.2k in edit train split. 

For each question-type, we finetune all 3 models with corresponding real + \vqaiv{} IQAs for the particular question type.  For a fair comparison, we also finetune all the models using just real data.  Figure \ref{fig:DA_ques_types12} (left) shows how different models, each specialized for a particular question type, behave when finetuned using real+synthetic data relative to the models finetuned using real data. The $y$ axis denoted the reduction in flips and $x$ axis represents the accuracy on the original set for each question type. We observe that using synthetic data always reduces flipping as all the points lie above the $y=0$ axis.  The amount of reduction in flips differs for each question type and varies from model to model. For instance, ques type `is this a' has the highest reduction in flips for CL model with no change in accuracy and has the least reduction of flips for ques type `how many'. However for SAAA, `how many' seems to have the highest reduction in flips with 2.5\% drop in accuracy. For SNMN, counting seems to have the highest reduction in flips. Moreover we also see that there are quite a few points on the right-hand side of $x=0$ axis which means that synthetic data also help improve accuracy on the test set.  Figure \ref{fig:DA_visual_aid} shows some of the examples for these specialized models. As we can see, finetuning the model with \vqaiv{} dataset helps in improving consistency and leads to more accurate predictions both on real as well as synthetic data.

Additionally, we also finetune all the baseline models with all the real data in VQA-v2 + \vqaiv{} data. Overall, we find that there is 5-6\% relative improvement in flips for all 3 models: CL (17.15$\rightarrow$16.1), SAAA (7.53$\rightarrow$7.09), SNMN (8.09$\rightarrow$7.72) with marginal improvement in accuracy\% in case of CL (60.21 $\rightarrow$60.24),  1\% reduction in accuracy in case of SAAA (70.25$\rightarrow$69.25) and 0.6\% improvement in accuracy for SNMN (67.65$\rightarrow$68.02). 

\myparagraph{CoVariant VQA Augmentation}
For counting, we create our \vqacv{} edit set by removing one instance of the object being counted and evaluate the models on both accuracy and consistency. Following the procedure above, we finetune all the models using real data, real+CV  and real+CV+IV IQAs. We evaluate the n/n-1 consistency for counting on \vqacv{} for all the three models. The results are shown in Figure \ref{fig:DA_ques_types12} (right). We see that using \vqacv{} edit set reduces flipping by 40\% for all 3 models with 1-4\% drop in accuracy. Additionally we see that using \vqacv{} + \vqaiv{} data reduce the flipping by 30\%: CL (83.8$\rightarrow$59.58), SAAA (77.74$\rightarrow$52.71), SNMN (77.13$\rightarrow$51.91)) with comparable accuracy: CL (43.65$\rightarrow$43.94), SAAA (50.87$\rightarrow$50.45) and SMNM (50.67$\rightarrow$50.61). Figure \ref{fig:DA_visual_aid} (Bottom) shows that models when trained using synthetic data can show a more accurate and consistent behaviour. Further consistency analysis with visualizations is in supplementary (section C.3).



\mario{don't say "more thorough" ... as this suggest that this analysis is not thorough. I've already changed.}

\section{Conclusion and Future Works}
We propose a semantic editing based approach to study and quantify the robustness of VQA models to visual variations. Our analysis shows that the models are brittle to visual variations and reveals spurious correlation being exploited by the models to predict the \VA{correct} answer. Next, we propose a data augmentation based technique to improve models' performance. Our trained models show significantly less flipping behaviour under invariant and covariant semantic edits, which we believe is an important step towards causal VQA models. Recent approaches to improve classifier robustness by regularizing them by exploiting data from different environments where only non-causal features vary~\cite{arjovsky2019invariant, heinze2018conditional}.
In our work, we explicitly create such data for the VQA task and make it available publicly, hoping it can support research towards building causal VQA models. 
\FloatBarrier

{\small
\bibliographystyle{ieee_fullname}
\bibliography{egbib}
}

\begin{appendices}
\section*{Appendix}
We structure the supplementary material as follows:
In Section A, we discuss further details about the human validation study. 
In Section B, we describe the VQA models used by us and mention the various hyperparameters used in section B.1. Following which are more visualizations showing predictions of the three VQA models on the original and the synthetic images from our proposed \vqaiv{}  and \vqacv{} datasets in section B.2. Also included in section B.2 are an analysis showing how the area of the removed object influences the flip rate and some attention maps for SAAA model on \vqaiv{} dataset.
Section C includes accuracy-flipping numbers for all the models finetuned using real vs real+edit IQAs for different question types for both \vqaiv{} and \vqacv{} datasets along with visualizations. Finally in Section D, we discuss a possible direction to introduce causality into VQA.

\section{Synthetic Dataset for Variances and Invariances in VQA}

\subsection{Human Validation}

In order to make sure that our consistency analysis holds and flipping is not due to errors in synthetic dataset, we collect all those IQAs for which labels flip (positively or negatively) for any of the three models (27621 IQAs, ~25\% of \vqaiv test set). Of this 25\%, we randomly sample 100 IQA from each of the 65 question categories \cite{vqa_v2} if possible. this results in a total of 4960 edited IQA.  Flipping of answers is bad and this number becomes our foundation for the robustness comparisons we make, so it was important for us to get this number validated. 
For each IQA, the annotator is asked to say if the answer shown is correct for the given image and question ( yes/no/ambiguous). We get these numbers validated by three humans and report the results in Table \ref{table:annotators_0.1_0.1_all_ans_same}. The study reveals that our edited IQA holds 91.3\% times according to all three humans. Additionally for 3.97\% IQAs: atleast one of them found it false and 5.68\% IQAs- seem to be ambiguous by atleast one of them.

\begin{table} 
\small
\centering
\begin{tabular}{ c  c c c  } 
 \toprule
  & Yes(\%) & No(\%) & Ambiguous(\%)\\ 
 \midrule
 User1  & 97.58 & 0.89 & 1.53 \\
 User2 & 96.47 & 1.15 & 2.38 \\
 User3 & 94.94 & 2.5 & 2.56 \\ 
 \midrule
 User1 $\cap$ User2 $\cap$ User3 & \textbf{91.31} & 0.04 & 0.04 \\ 
 User1 $\cup$ User2 $\cup$ User3 & 99.6 & \textbf{3.87} & \textbf{5.68} \\ [1ex]
 \bottomrule

\end{tabular}
\caption{Human Validation of the edited set: If the given answer is valid for the Image-Question pair. }
\label{table:annotators_0.1_0.1_all_ans_same}
\end{table}

\section{Experiments: Consistency analysis}

\subsection{Models Training}
We select three models for our comparison. The first one is a basic CNN+LSTM model, where we use ResNet152 \cite{ResNet} pre-trained on Image-Net \cite{imagenet_cvpr09}, to embed the images. The question features are obtained by feeding the tokenized and encoded input question embeddings into the LSTM. The features are then concatenated and fed to the classifier to infer the answer. 
Secondly, we use an attention model- Show, Ask, Attend and Answer (SAAA) described by Kazemi \emph{et~al.} in \cite{SAAA}. The aim of the attention models is the identify and use local image features with different weights. After processing images using ResNet and feeding tokenized questions to LSTM, the concatenated image features and the final state of LSTMs are used to compute multiple attention distributions over image features.
Lastly, we use a compositional model, SNMN \cite{SNMN}. The model consists of three different components: layout controller to decompose the question into a sequence of sub-tasks, set of neural modules to perform the sub-tasks and a differentiable memory stack.

For training these models, we use the codes available online with the specified hyperparamters. For SNMN, we use official code available to train the model, \cite{github_snmn}. For training SAAA, we use the code available online, \cite{pytorch_vqa_SAAA}. We modified the available SNMN code in order to get CNN+LSTM model- we just removed the attention layers from the network. As we use the validation split for consistency evaluation and testing, we cannot let the models train on it. We keep aside the validation set for testing, and only the training split is used to train the models. All these models use standard Cross Entropy Loss and follow the standard VQA practices. 
We follow all respective pre-processing  and training procedure given on the github sites (SNMN: link \cite{github_snmn},SAAA link: \cite{github_snmn} for pre-processing IQAs and for training. For SAAA and CNN+LSTM: ResNet152, conv layer-4 is used to extract $14*14$ features for image whereas SNMN uses ResNet 152, layer-5 resulting into $7*7$ features. The learning rate used to train each model is $e^{-3}$, batch size for learning is set to 128 for all 3 models.

\subsection{Visualizations}

\begin{figure*}
\begin{center}
\includegraphics[width=1\linewidth]{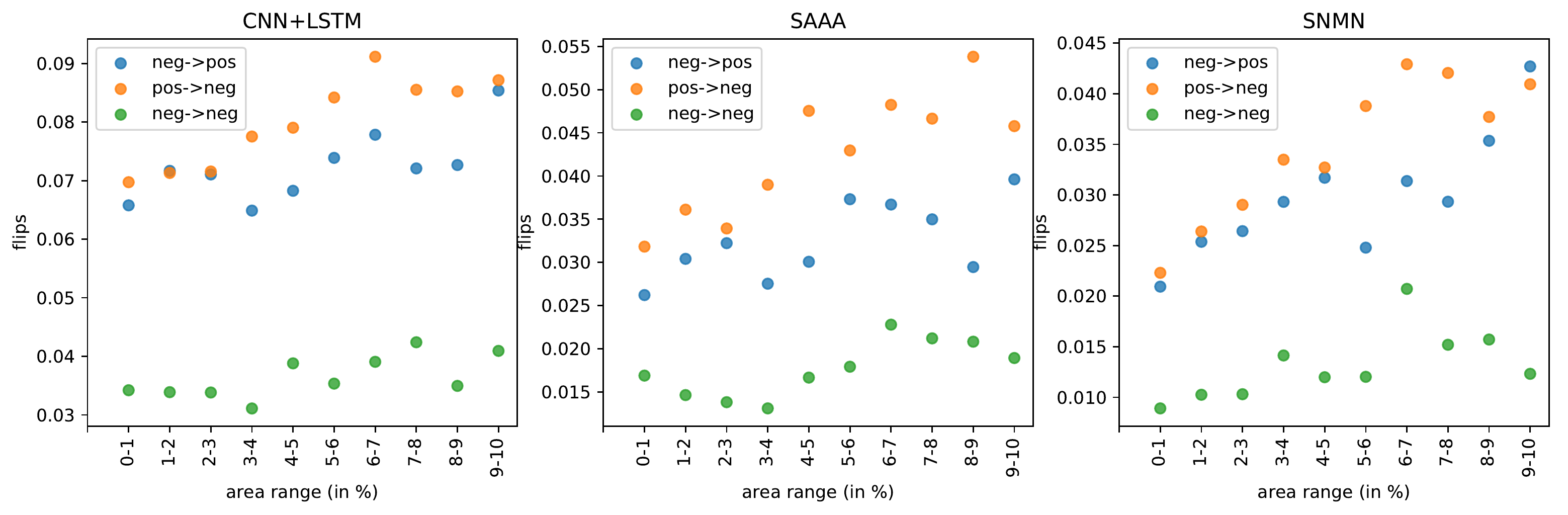}
\end{center}
\caption{Flip rate vs. area of the object.}
\label{fig:area_flips_plot}
\end{figure*}

\begin{figure*}
\small
\centering
  \begin{tabular}{l c c l c c }
    \toprule
    \multicolumn{3}{c}{ Are there any grapes? }   & \multicolumn{3}{c}{ What room of a house is this? }   \\
    
    \multicolumn{2}{c}{\tab[1.5cm] \myit{A}: yes} & \multicolumn{1}{c}{\tab[0.2cm] \textit{banana removed}; \myit{A}: yes}  & \multicolumn{2}{c}{\tab[1.5cm]\myit{A}: kitchen} & \multicolumn{1}{c}{\tab[0.2cm]\textit{bowl removed}; \myit{A}: kitchen}  \\
    
    \multicolumn{3}{c}{\includegraphics[width=0.5\textwidth]{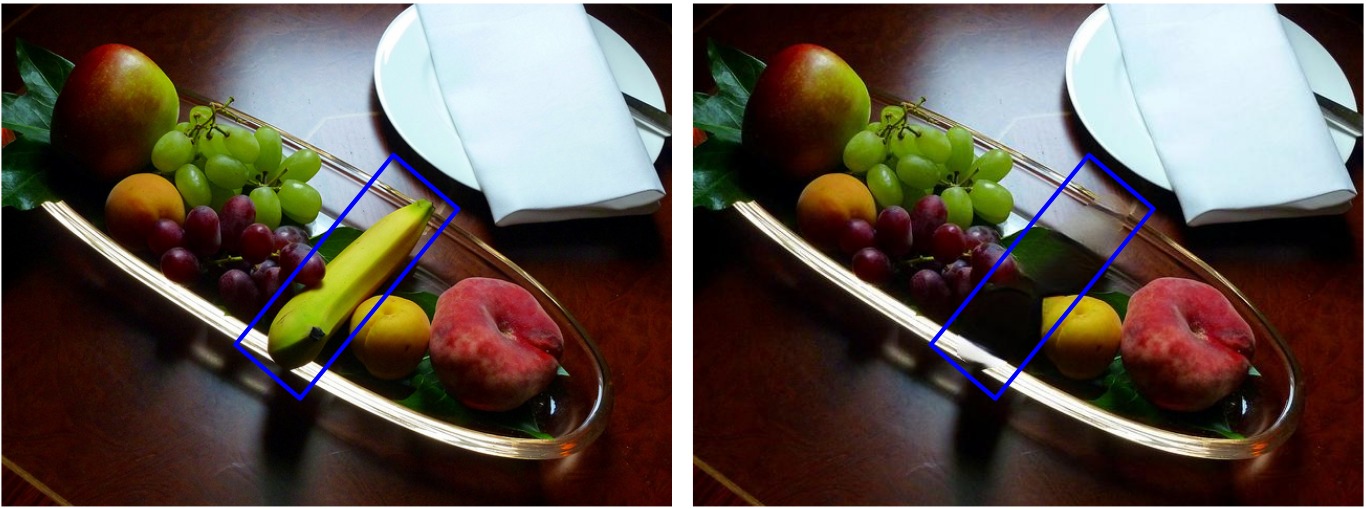}} & \multicolumn{3}{c}{\includegraphics[width=0.5\textwidth]{Figures/vase_bathroom_kitchen}}\\

    SAAA & \red{no} & \red{no} & SAAA & \red{bathroom}  & \green{kitchen} \\

    \multicolumn{3}{c}{\includegraphics[width=0.25\textwidth,trim= 2 2 2 2, clip]{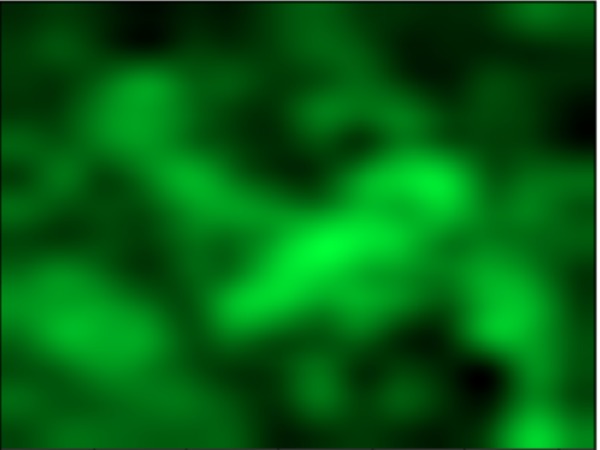} \includegraphics[width=0.25\textwidth,trim= 2 2 2 2, clip]{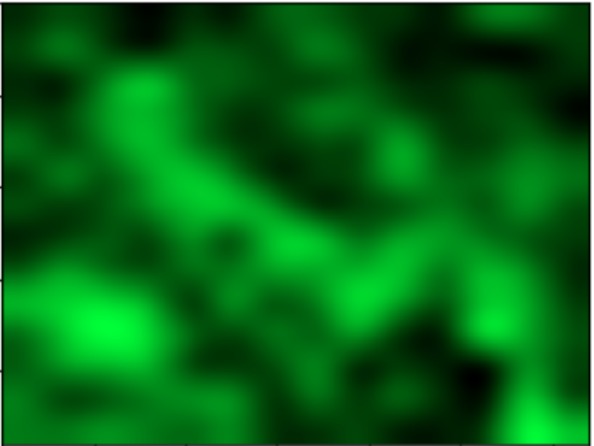}} &     \multicolumn{3}{c}{\includegraphics[width=0.25\textwidth,trim= 2 2 2 2, clip]{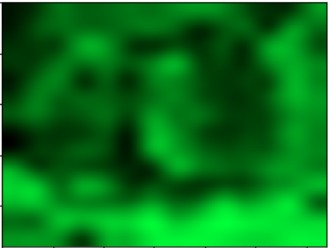} \includegraphics[width=0.25\textwidth,trim= 2 2 2 2, clip]{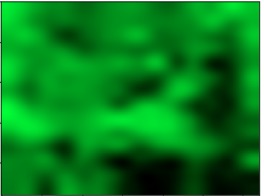}}\\
        
    \multicolumn{3}{c}{\includegraphics[width=0.25\textwidth,trim= 2 2 2 9, clip ]{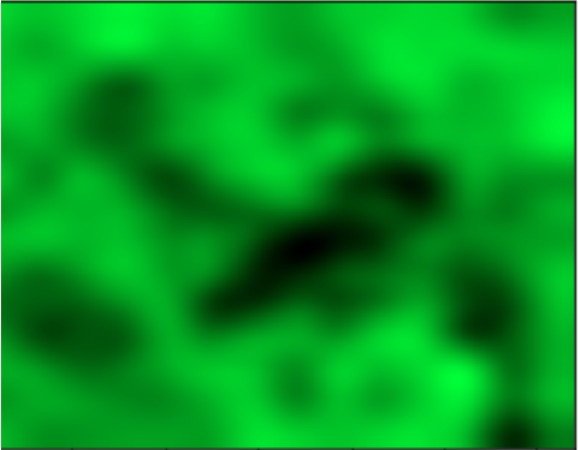} \includegraphics[width=0.25\textwidth,trim= 2 2 2 1, clip]{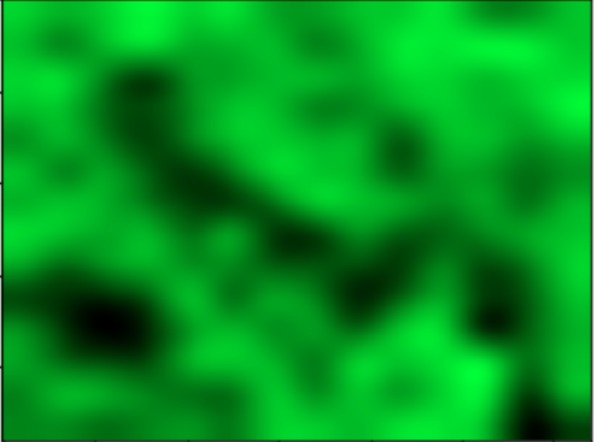}}&
    \multicolumn{3}{c}{\includegraphics[width=0.25\textwidth,trim= 2 2 2 2, clip]{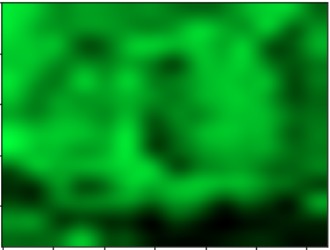} \includegraphics[width=0.25\textwidth,trim= 2 2 2 2, clip]{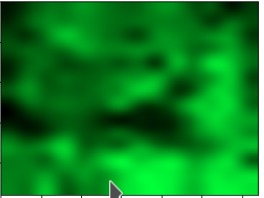}}\\ 
    
    \toprule

  \end{tabular}
\caption{Shown above are the attention maps for SAAA on original and edited images from our synthetic dataset \vqaiv. The attention maps are diffuse and does not clearly show one object where the model pays attention to.}
\label{fig:heatmap_visual}
\end{figure*}

\begin{figure*}
\small
\centering
  \begin{tabular}{l c c l c c }
    \toprule
    
    \multicolumn{3}{c}{Q: What color is the bird house?} & \multicolumn{3}{c}{Q: What color is the sauce? } \\ 
    
    \multicolumn{2}{c}{\myit{A}: yellow} & \multicolumn{1}{r}{\textit{baseball bat removed}; \myit{A}: yellow}
    &     \multicolumn{2}{c}{\myit{A}: red} & \multicolumn{1}{c}{\textit{cup removed}; \myit{A}: red}\\
    
    \multicolumn{3}{c}{\includegraphics[width=0.5\textwidth,trim= 0 0 0 0, clip]{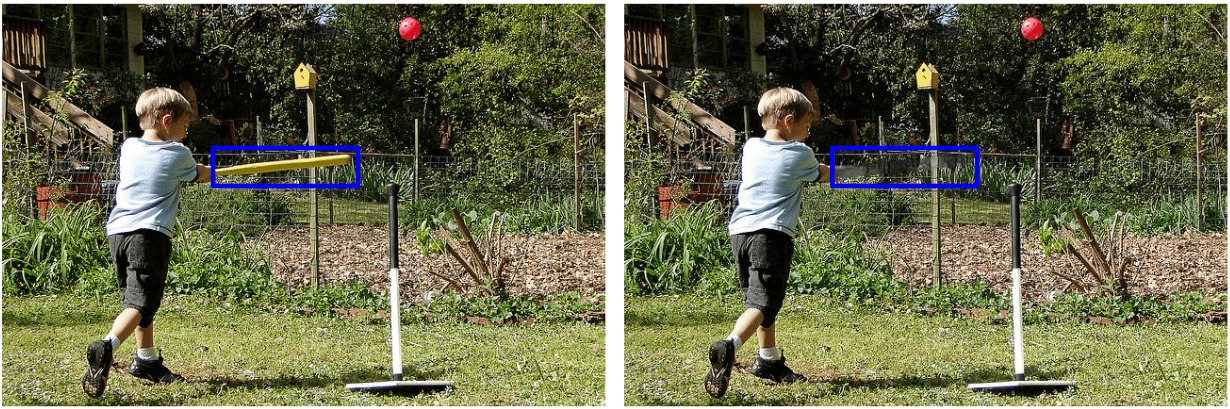}} &     \multicolumn{3}{c}{\includegraphics[width=0.5\textwidth, trim= 0 0 0 0, clip]{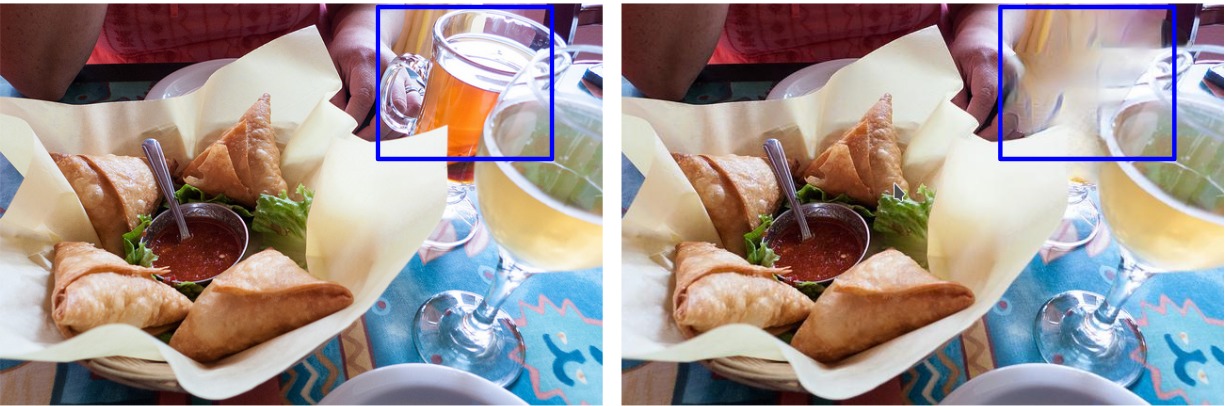}}\\
    CNN+LSTM & \green{yellow} & \red{red} &    CNN\_LSTM & \green{red} &   \red{white}  \\
    SAAA & \red{white } &  \red{white} & SAAA  & \red{orange} &    \green{red}   \\ 
    SNMN & \green{yellow } & \textcolor{red}{white}  & SNMN &  \red{orange} &  \red{yellow}   \\ 

    \toprule
    
    \multicolumn{3}{c}{Q: What color is the toilet seat?} &    \multicolumn{3}{c}{Q: What color is cone?} \\ 
    
    \multicolumn{2}{c}{\myit{A}: white} & \multicolumn{1}{c}{\textit{sink removed}; \myit{A}: white}
    &     \multicolumn{2}{c}{\myit{A}: orange} & \multicolumn{1}{c}{\textit{person removed}; \myit{A}: orange}\\
    
    \multicolumn{3}{c}{\includegraphics[width=0.5\textwidth, trim= 0 0 0 160, clip]{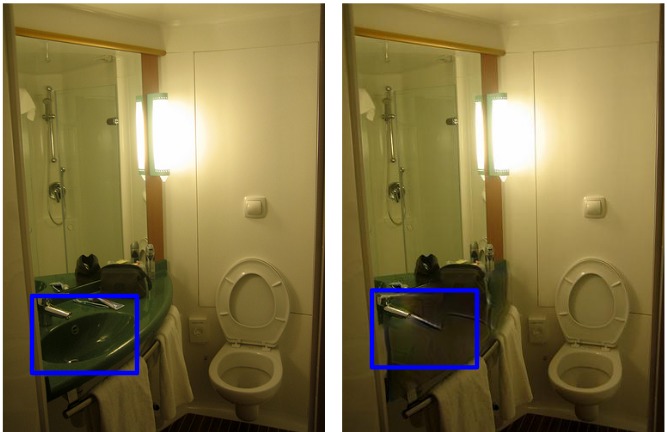}} &   \multicolumn{3}{c}{\includegraphics[width=0.5\textwidth, trim= 0 0 0 0, clip]{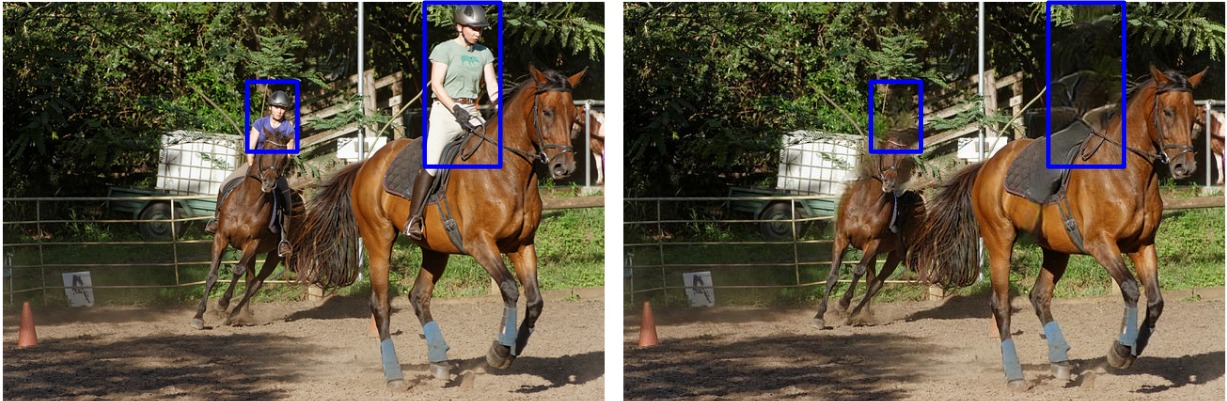}}\\
    \Xhline{2\arrayrulewidth}
    
    CNN\_LSTM & \red{green} &    \green{white} &     CNN\_LSTM & \red{brown} &  \green{orange} \\
    SAAA  & \red{green} &    \red{brown}  &     SAAA  & \green{orange} &  \red{white}   \\ 
    SNMN &  \red{green} &    \red{brown}   &     SNMN &  \red{white} &   \green{orange}   \\
    
    \toprule

    \multicolumn{3}{c}{ Is this a kite?} &  \multicolumn{3}{c}{Q: Is this a museum? } \\
    
    \multicolumn{2}{c}{\myit{A}: yes} & \multicolumn{1}{c}{\textit{backpack removed}; \myit{A}: yes}  & \multicolumn{2}{c}{\myit{A}: no} & \multicolumn{1}{c}{\textit{couch removed}; \myit{A}: no}  \\
        
    \multicolumn{3}{c}{\includegraphics[width=0.5\textwidth, trim= 0 20 0 0, clip]{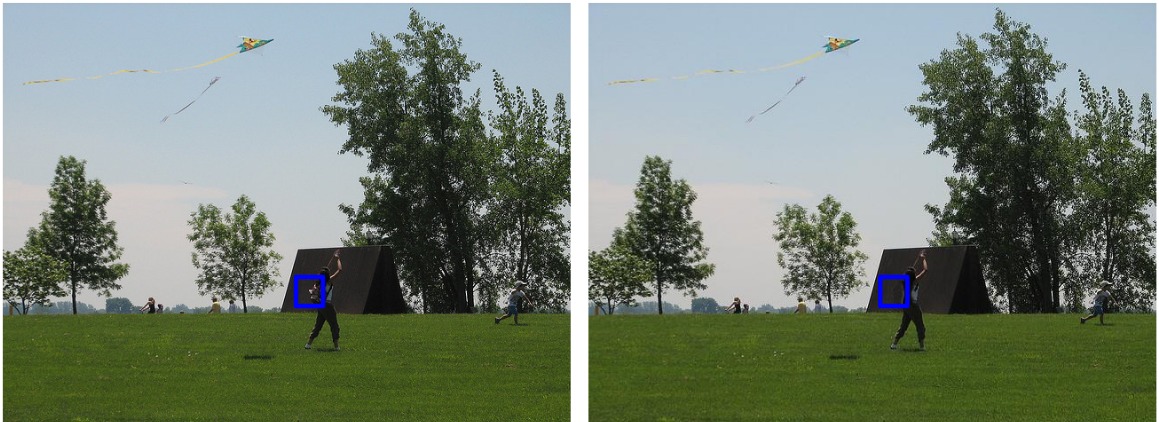}} & 
    \multicolumn{3}{c}{\includegraphics[width=0.5\textwidth, trim= 0 0 0 0, clip]{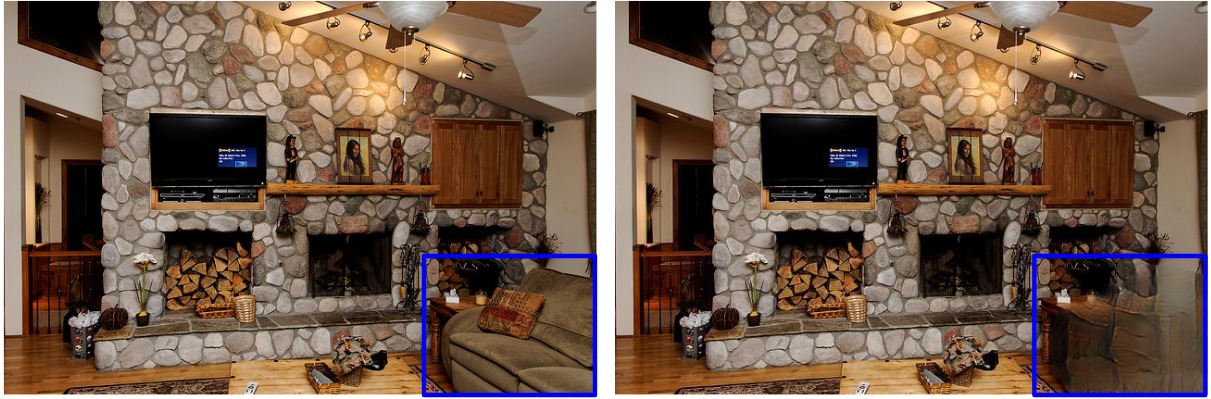}}\\
    
     CNN\_LSTM & \green{yes}   & \red{no}   & CNN\_LSTM & \green{no}  & \red{yes}  \\
     SAAA  & \red{no}  & \red{no}   & SAAA  & \red{yes}  & \red{yes}   \\  
     SNMN &  \green{yes}   & \red{no}  & SNMN &  \green{no}  & \red{yes}   \\
 
    \toprule

    \multicolumn{3}{c}{ How many bowls of food are there? } & \multicolumn{3}{c}{ How many desk lamps are there? }   \\
    
    \multicolumn{2}{c}{\myit{A}: 2} & \multicolumn{1}{c}{\textit{bottle removed}; \myit{A}: 2}  & \multicolumn{2}{c}{\myit{A}: 1} & \multicolumn{1}{c}{\textit{laptop removed}; \myit{A}: 1}  \\
        
    \multicolumn{3}{c}{\includegraphics[width=0.5\textwidth]{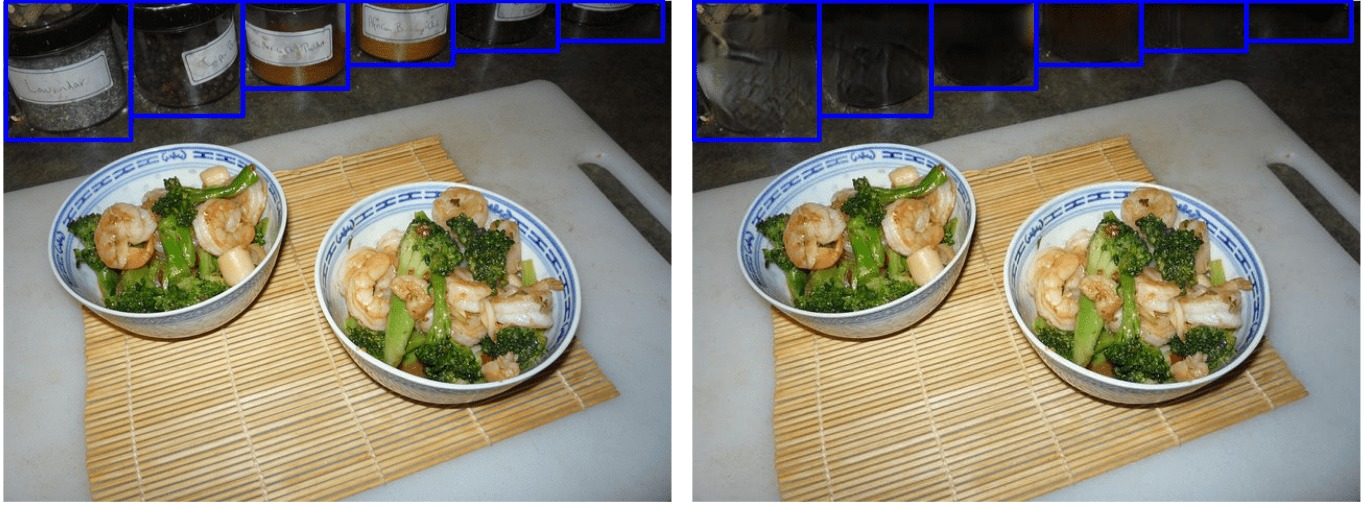}} & \multicolumn{3}{c}{\includegraphics[width=0.5\textwidth]{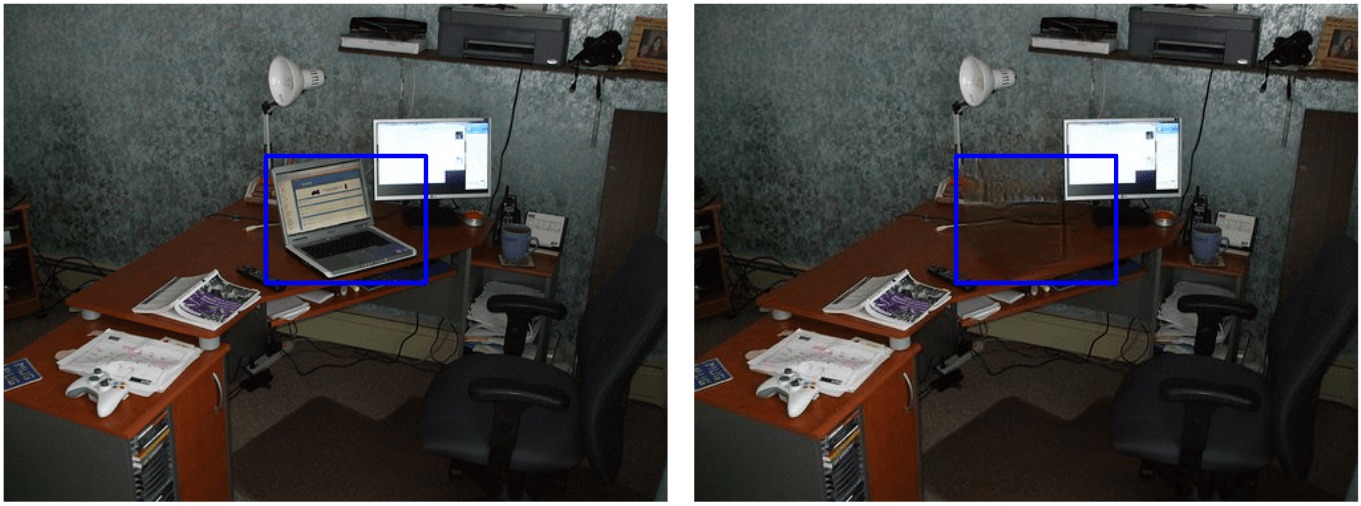}}\\

    CNN+LSTM & \red{3 } & \red{3} & CNN+LSTM & \red{2} & \green{1} \\
    SAAA & \red{3 } & \green{2 } & SAAA & \red{0} & \green{1}  \\
    SNMN & \green{2 } & \red{1 } &  SNMN & \green{1} & \green{1} \\

    \toprule

  \end{tabular}
\caption{Models tend to look at different objects while predicting the answers. Shown above are the models' predictions on original and edited images from our synthetic dataset \vqaiv. }
\label{fig:vqa_iv_flip1}
\end{figure*}

\begin{figure*}
\centering
\small
  \begin{tabular}{l c c l c c}
    
    \toprule

    \multicolumn{3}{c}{ Are there any grapes? }   & \multicolumn{3}{c}{ Is there a trash can? }   \\
    
    \multicolumn{2}{c}{\myit{A}: yes} & \multicolumn{1}{c}{\textit{banana removed}; \myit{A}: yes}  & \multicolumn{2}{c}{\myit{A}: yes} & \multicolumn{1}{c}{\textit{toilet removed}; \myit{A}: yes}  \\
    
    \multicolumn{3}{c}{\includegraphics[width=0.5\textwidth]{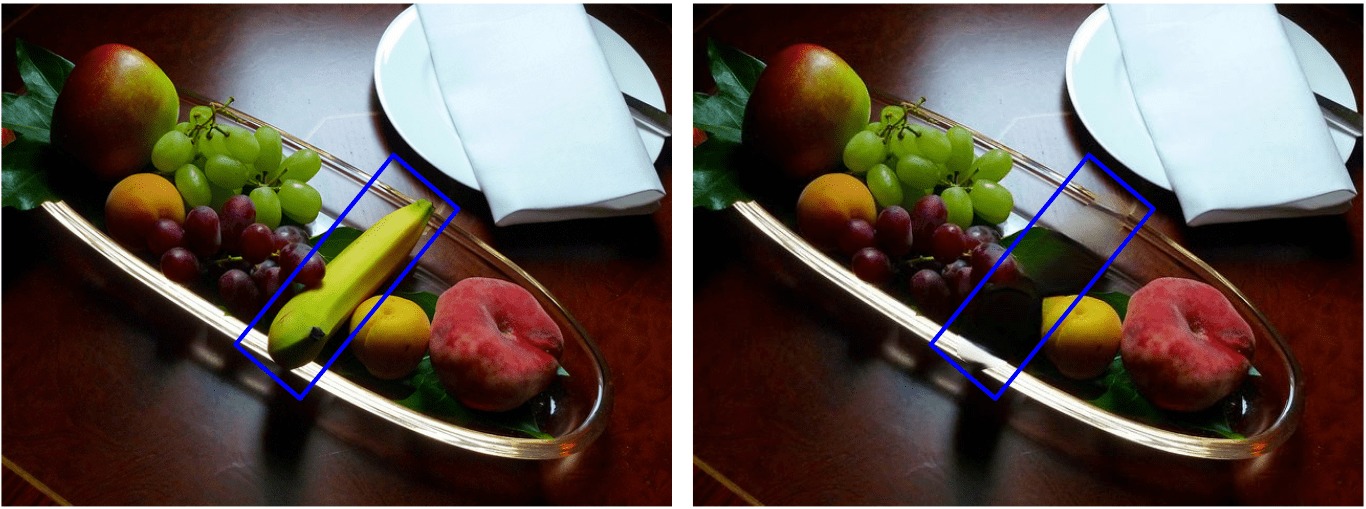}} & \multicolumn{3}{c}{\includegraphics[width=0.5\textwidth]{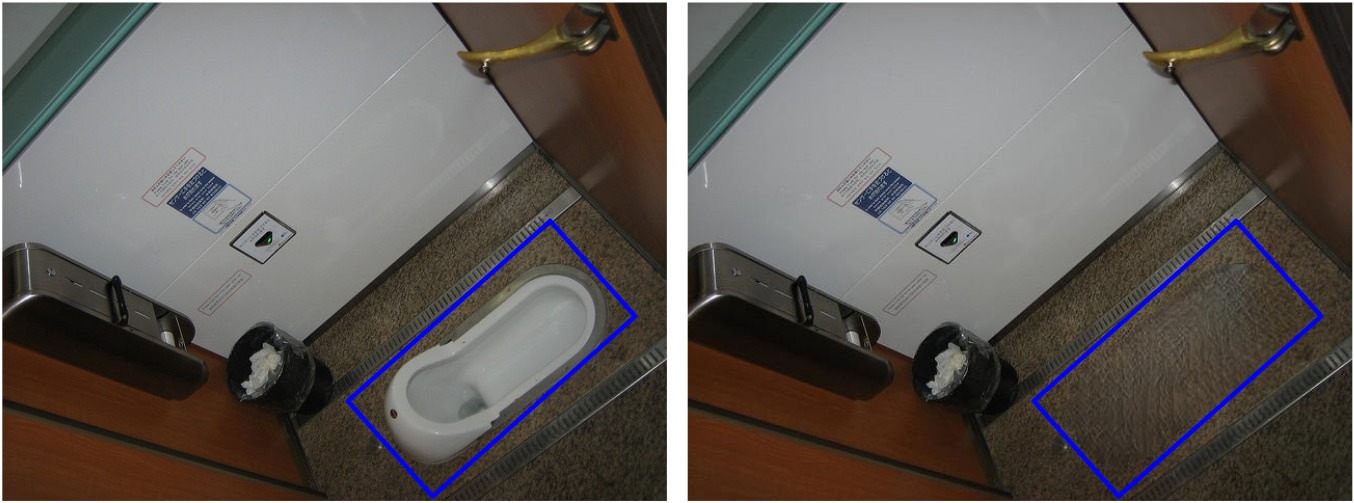}}\\

    CNN+LSTM & \green{yes} & \red{no}  & CNN+LSTM & \red{no} & \green{yes} \\
    SAAA & \red{no} & \red{no} & SAAA & \green{yes} & \green{yes} \\
    SNMN & \red{no} & \green{yes} & SNMN & \red{no} & \green{yes} \\

    \toprule

    \multicolumn{3}{c}{ What is the liquid in the pitcher?} & \multicolumn{3}{c}{ Is there an airport nearby? [car] }\\

    \multicolumn{2}{c}{\myit{A}: water} & \multicolumn{1}{c}{\textit{wine glass removed}; \myit{A}: water}  & \multicolumn{2}{c}{\myit{A}: yes} & \multicolumn{1}{c}{\textit{car removed}; \myit{A}: yes}  \\
    
    \multicolumn{3}{c}{\includegraphics[width=0.5\textwidth, trim= 0 20 0 10, clip]{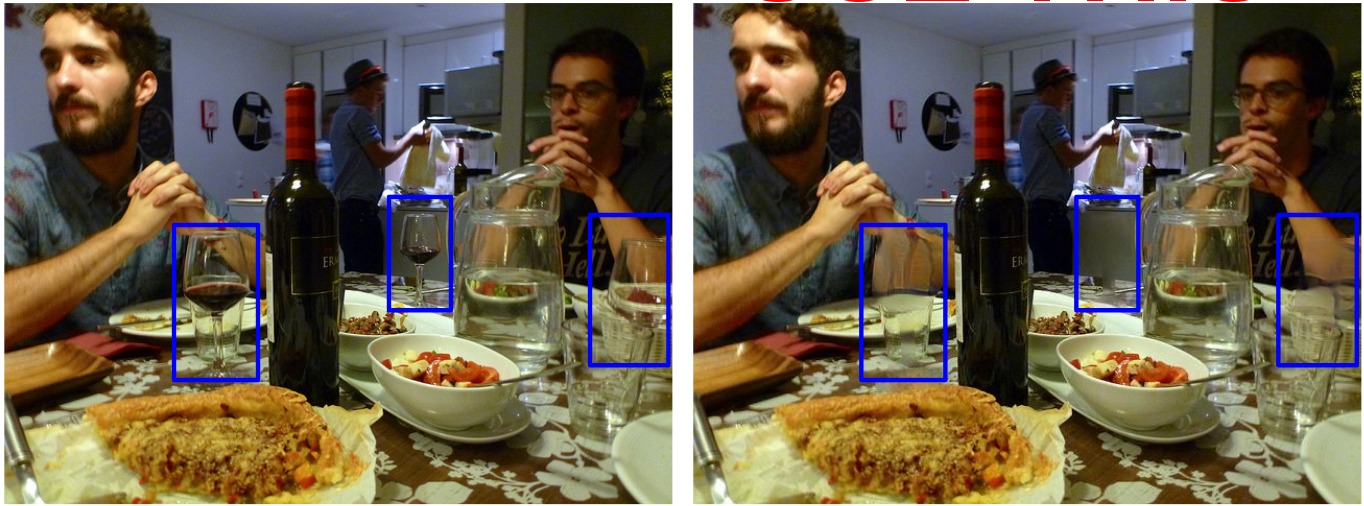}} &     \multicolumn{3}{c}{\includegraphics[width=0.5\textwidth, trim= 0 0 0 0, clip]{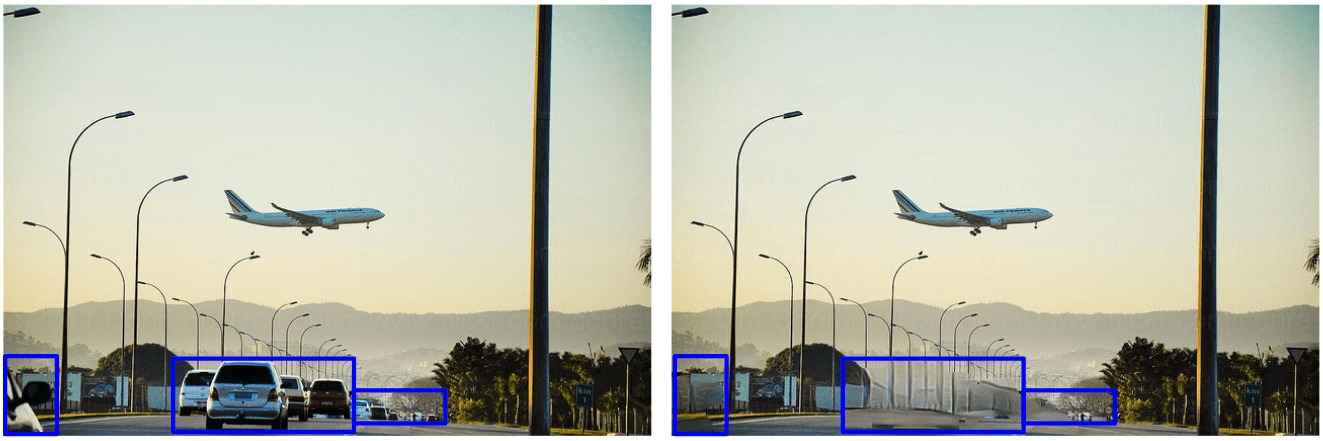}} \\
        
    CNN+LSTM & \red{wine} & \red{beer} &  CNN+LSTM & \green{yes} & \red{no}  \\ 
    SAAA & \red{wine} & \red{wine} & SAAA  & \green{yes} & \red{no}   \\ 
    SNMN & \red{wine} & \green{water}  & SNMN  & \green{yes} & \red{no} \\

    \toprule

    \multicolumn{3}{c}{ What is in the sky?} & \multicolumn{3}{c}{ What is room to the right called?}  \\

    \multicolumn{2}{c}{\myit{A}: nothing} & \multicolumn{1}{c}{\textit{airplane removed}; \myit{A}: nothing}  & \multicolumn{2}{c}{\myit{A}: kitchen} & \multicolumn{1}{c}{\textit{toilet removed}; \myit{A}: kitchen}  \\
        
    \multicolumn{3}{c}{\includegraphics[width=0.5\textwidth, trim= 0 25 0 0, clip]{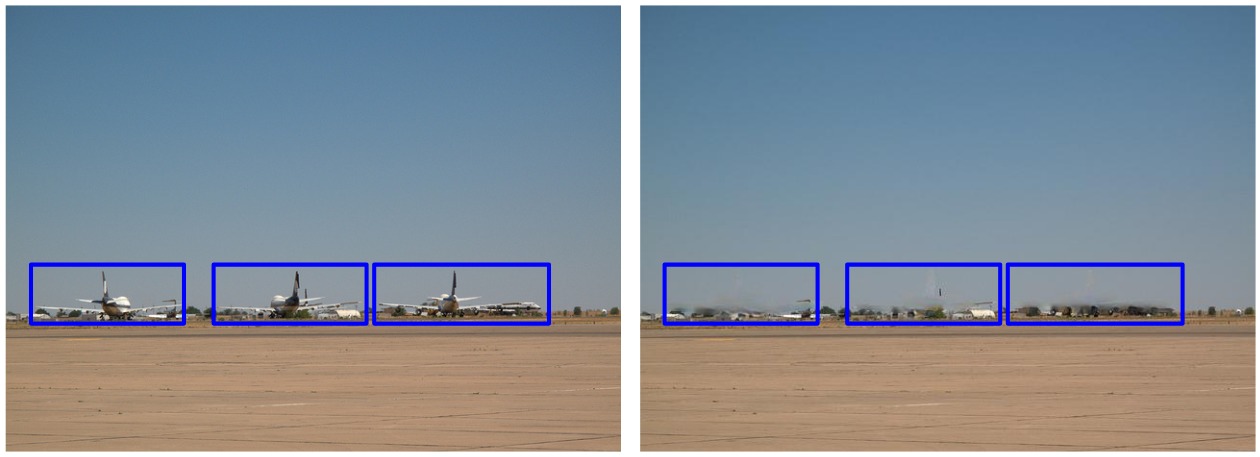}} &     \multicolumn{3}{c}{\includegraphics[width=0.5\textwidth, trim= 0 0 0 0, clip]{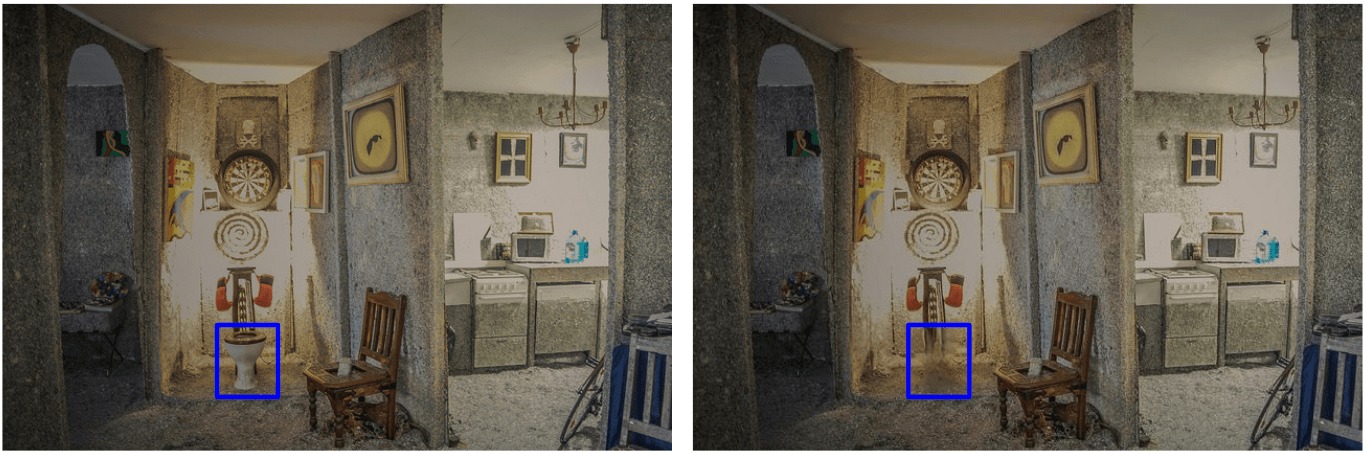}}\\
    
    CNN+LSTM & \red{plane} & \red{kite}  &     CNN+LSTM & \red{bathroom} & \red{living room} \\
    SAAA & \red{plane} &  \red{kite}  & SAAA & \red{bathroom} &  \red{living room}  \\
    SNMN & \red{plane} & \red{clouds}  & SNMN & \red{bathroom} & \red{living room}  \\
    \Xhline{2\arrayrulewidth}
    
    \multicolumn{3}{c}{ What is the purple thing? } & \multicolumn{3}{c}{ What are the kids doing? } \\

    \multicolumn{2}{c}{\myit{A}: pillow} & \multicolumn{1}{c}{\textit{remote removed}; \myit{A}: pillow}  & \multicolumn{2}{c}{\myit{A}: petting horse} & \multicolumn{1}{c}{\textit{bench removed}; \myit{A}: petting horse}  \\
     
    \multicolumn{3}{c}{\includegraphics[height=0.17\textwidth, width=0.5\textwidth, trim= 0 40 0 80, clip]{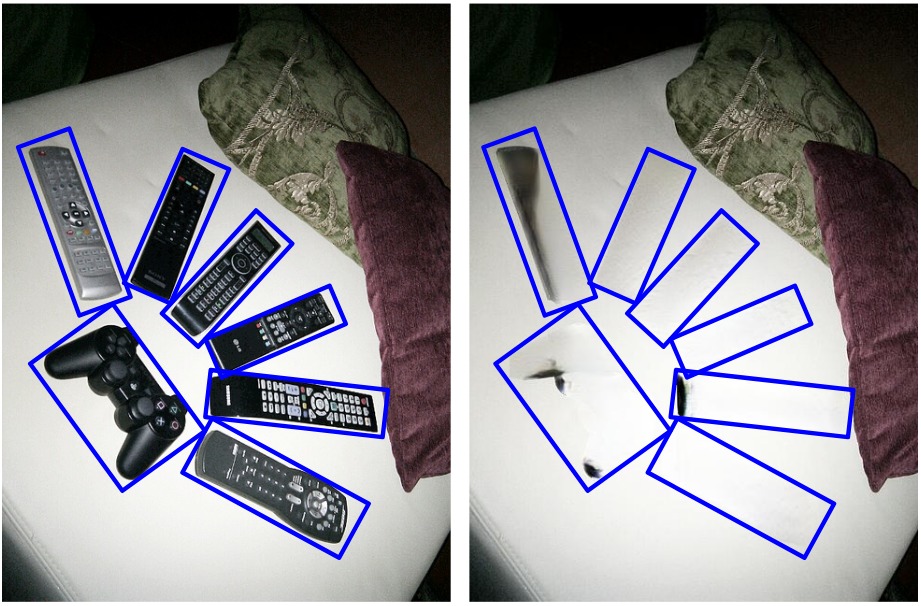}} &     \multicolumn{3}{c}{\includegraphics[width=0.5\textwidth, trim= 0 0 0 0, clip]{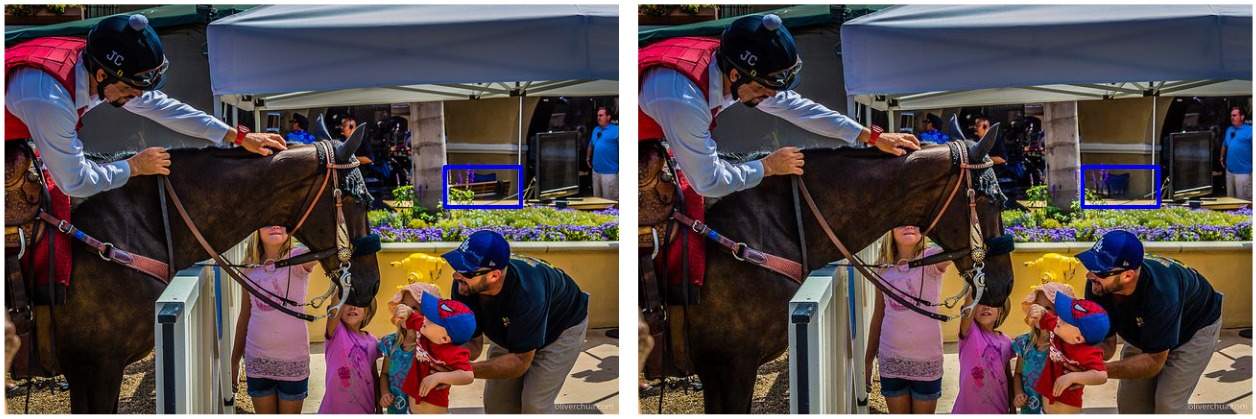}}\\
    CNN+LSTM & \red{scissors} & \red{heart} & CNN+LSTM & \red{racing} & \red{riding horse} \\
    SAAA & \red{remote} & \red{blanket} & SAAA & \red{standing} & \red{shaking hands} \\
    SNMN & \red{remote} & \red{blanket} & SNMN & \red{playing} & \red{racing} \\

    \Xhline{2\arrayrulewidth}
    
  \end{tabular}
  \caption{Existing VQA models are brittle are brittle to semantic variations in the images. Shown above are examples showing different sorts of flips for \vqaiv}
  \label{fig:vqaiv_flip2}
\end{figure*}

Figures \ref{fig:vqa_iv_flip1}, \ref{fig:vqaiv_flip2} show the predictions of 3 models on original and edited IQA from \vqaiv dataset. We expect the models to make consistent predictions across original and edited images. However we see that this isn't the case.  

\begin{figure*}
\centering
\small
  \begin{tabular}{l c c l c c}
    \toprule

    \multicolumn{3}{c}{ How many animals? } & \multicolumn{3}{c}{ How many planes are in the air? } \\
    
    \multicolumn{2}{c}{\myit{A}: 1} & \multicolumn{1}{c}{\tab[2cm] \textit{giraffe removed}; \myit{A}: 0}  & \multicolumn{2}{c}{\myit{A}: 1} & \multicolumn{1}{c}{\tab[2cm] \textit{plane removed}; \myit{A}: 0}  \\
        
    \multicolumn{3}{c}{\includegraphics[width=0.48\textwidth, trim= 0 0 0 0, clip]{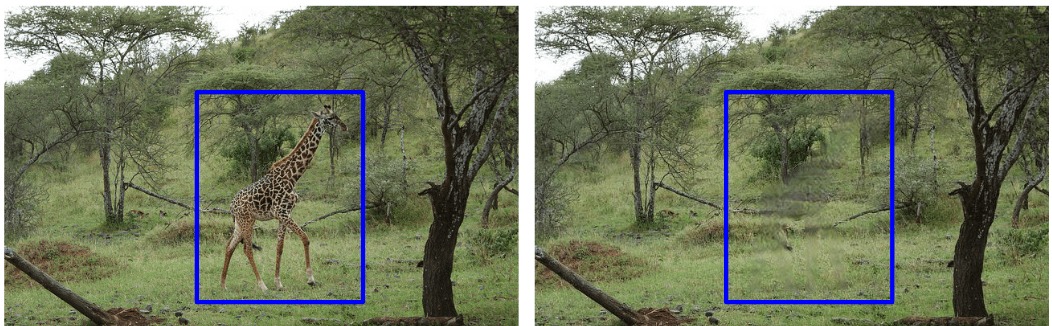}} &
    \multicolumn{3}{c}{\includegraphics[width=0.48\textwidth, trim= 0 20 0 30, clip]{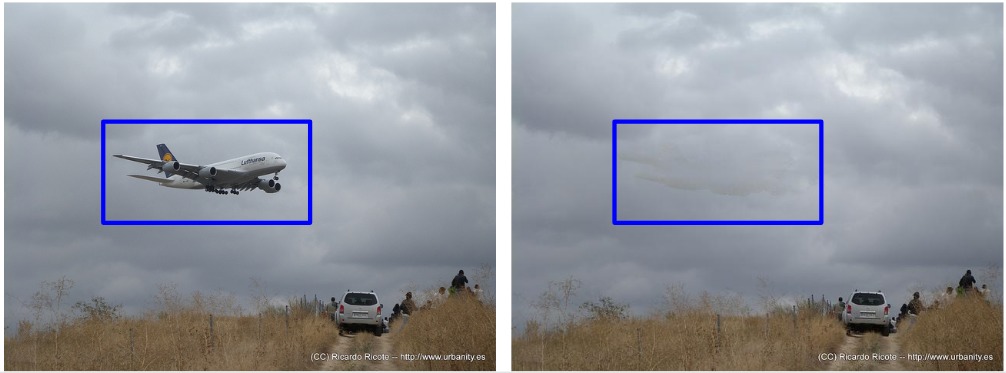}}\\
    CNN\_LSTM & \green{1}  & \red{1}  & CNN\_LSTM & \green{1}  & \red{1} \\
    SAAA  & \green{1}   & \red{1}  & SAAA  & \green{1}  & \red{1} \\  
    SNMN &  \green{1}   & \red{1}   & SNMN & \green{1}  & \red{1}   \\
    
    \toprule

    \multicolumn{3}{c}{ How many clocks are there? } & \multicolumn{3}{c}{ How many children are there? } \\
    
    \multicolumn{2}{c}{\myit{A}: 2} & \multicolumn{1}{c}{\tab[2cm] \textit{clock removed}; \myit{A}: 1}  & \multicolumn{2}{c}{\myit{A}: 5} & \multicolumn{1}{c}{\tab[2cm] \textit{child removed}; \myit{A}: 4}  \\
        
    \multicolumn{3}{c}{\includegraphics[width=0.48\textwidth, trim= 0 0 0 0, clip]{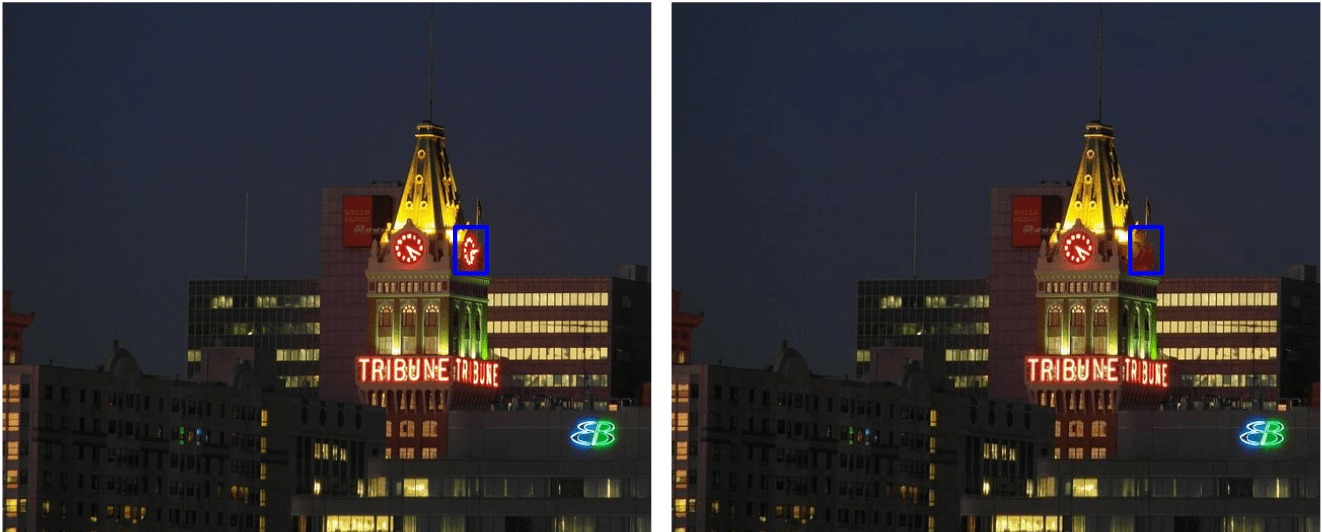}} &
    \multicolumn{3}{c}{\includegraphics[width=0.48\textwidth, trim= 0 80 0 0, clip]{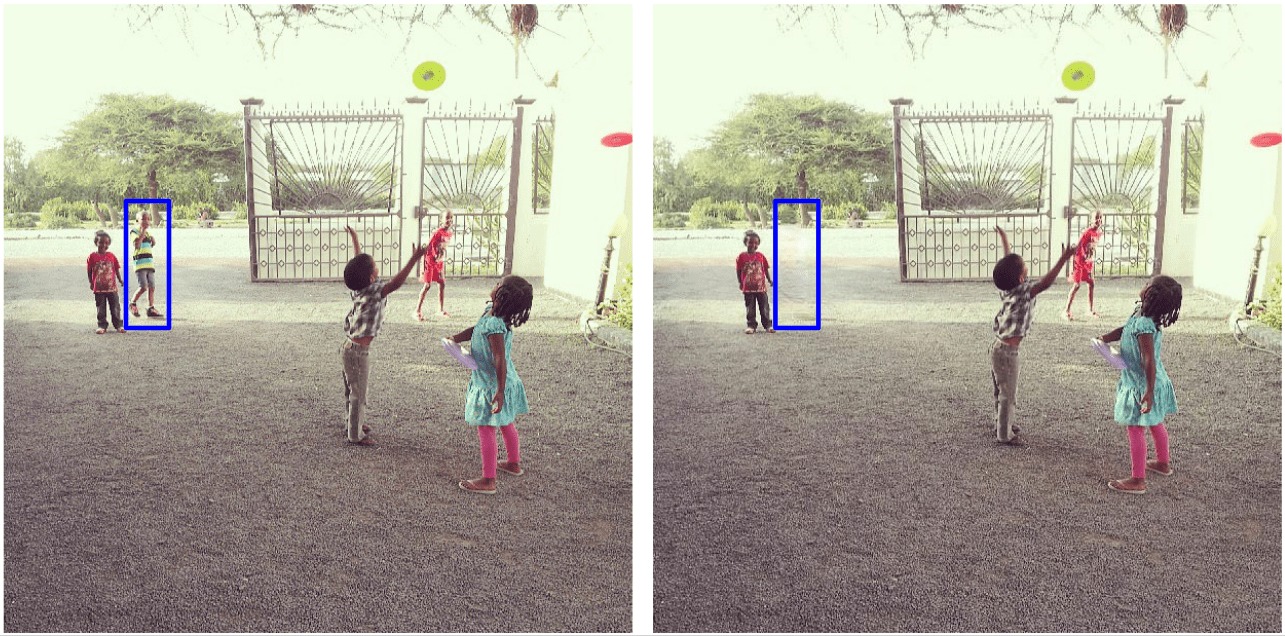}}\\
    CNN+LSTM & \red{2} & \green{1} & CNN+LSTM & \red{2} & \red{2} \\
    SAAA & \red{2} & \green{1} & SAAA & \red{2} & \red{2} \\
    SNMN & \red{2} & \green{1} & SNMN & \red{2} & \red{2} \\
    
    \toprule

    \multicolumn{3}{c}{ How many horses are in the picture?} & \multicolumn{3}{c}{ How many zebras? } \\
    
    \multicolumn{2}{c}{\myit{A}: 1} & \multicolumn{1}{c}{\tab[2cm] \textit{horse removed}; \myit{A}: 0}  & \multicolumn{2}{c}{\myit{A}: 2} & \multicolumn{1}{c}{\tab[2cm] \textit{zebra removed}; \myit{A}: 1}  \\
        
    \multicolumn{3}{c}{\includegraphics[width=0.48\textwidth, trim= 0 0 0 0, clip]{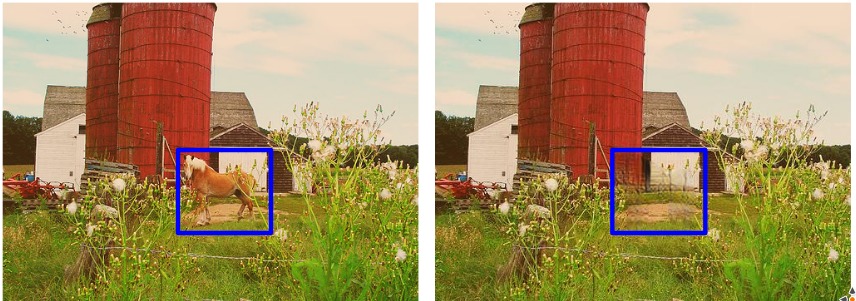}} &
    \multicolumn{3}{c}{\includegraphics[width=0.48\textwidth, trim= 0 0 0 0, clip]{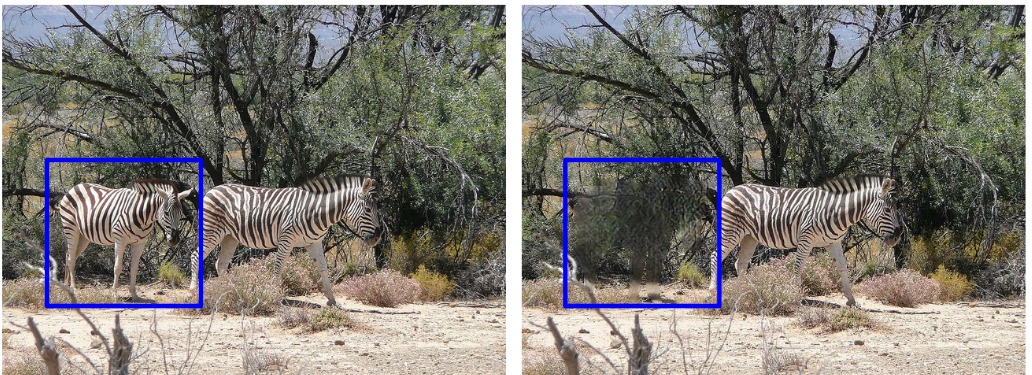}}\\
    CNN\_LSTM & \red{2}  & \red{1}  & CNN\_LSTM & \red{3}   & \red{2} \\
    SAAA  & \red{2}   & \red{1}  & SAAA  & \red{3}  & \red{2}  \\  
    SNMN &  \green{1}   & \red{1}   & SNMN &  \green{2}  & \red{2}   \\
    
    \toprule

    \multicolumn{3}{c}{ How many people are in the image?} & \multicolumn{3}{c}{ How many giraffes are here?} \\
    
    \multicolumn{2}{c}{\myit{A}: 1} & \multicolumn{1}{c}{\tab[2cm] \textit{person removed}; \myit{A}: 0}  & \multicolumn{2}{c}{\myit{A}: 3} & \multicolumn{1}{c}{\tab[2cm] \textit{giraffe removed}; \myit{A}: 2}  \\
        
    \multicolumn{3}{c}{\includegraphics[width=0.48\textwidth, trim= 0 0 0 0, clip]{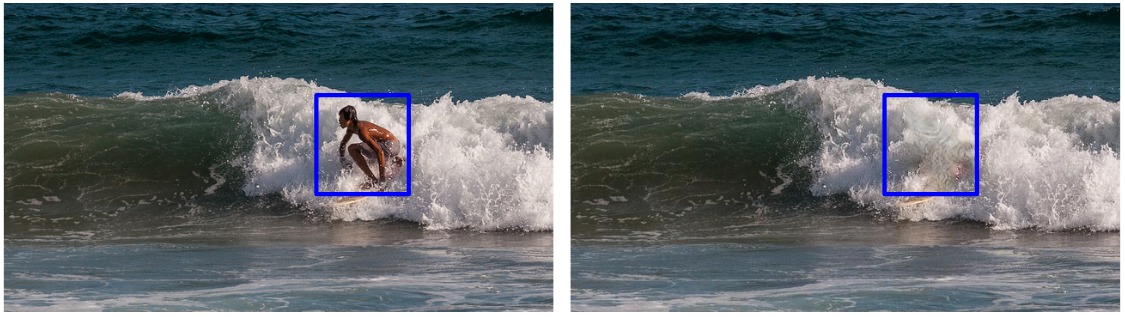}} &
    \multicolumn{3}{c}{\includegraphics[width=0.48\textwidth, trim= 0 10 0 30, clip]{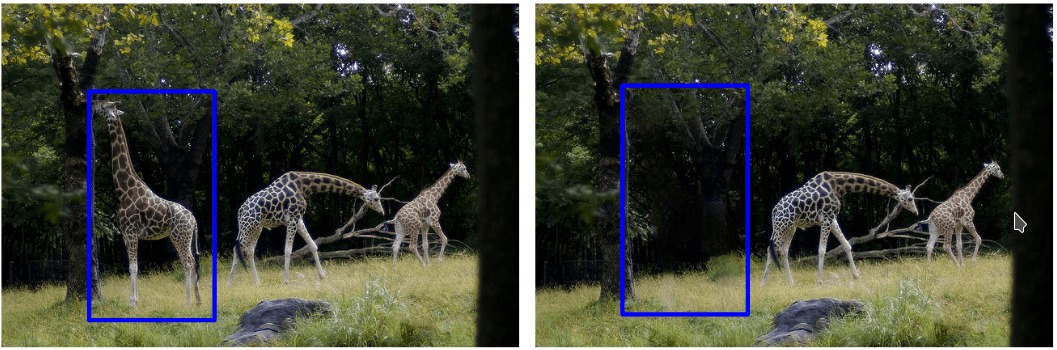}}\\
    
    CNN\_LSTM & \green{1}  & \red{1}  & CNN\_LSTM & \red{2}  & \red{1} \\
    SAAA  & \green{1}   & \red{1}  & SAAA  & \red{2}  & \green{2} \\  
    SNMN &  \green{1}   & \red{1}   & SNMN & \green{3}  & \green{2}   \\
    
    \toprule
    
  \end{tabular}
  \caption{Shown above are models' predictions on original and edited images from \vqacv. }
  \label{fig:vqacv_flip}
\end{figure*}

Figure \ref{fig:vqacv_flip} shows the predictions on original and edited IQA from \vqacv dataset. Here we expect the models to maintain n/n-1 consistency. Counting is a hard problem for VQA models and enforcing consistency seems to break these models completely. 

\textbf{Area of the object removed vs flip.}
 To study the correlation of area of the object removed on different types of flips, we plot the flip rate for different area ranges for objects being removed in Figure \ref{fig:area_flips_plot}. According to our analysis, there is no large dependence between removed object area and the flip rate. For example, for objects of size 0-1\% of the image area, {pos$\rightarrow$neg}  flip rate was about 7\% for CNN-LSTM and for objects of size 9-10\%, the flip rate only marginally higher at 8.7\%.

\textbf{Heatmaps for inspection.}
SAAA\cite{SAAA} has attention mechanisms incorporated in its architecture. The model uses concatenated image features and final state of LSTMs to
compute multiple attention distributions over image features. One would expect these attention maps to provide a clue as to where the model is looking in order to explain the flipping behaviour under editing. To see if this is true we visualize the attention maps for SAAA  on original/edited examples in Figure \ref{fig:heatmap_visual}.  On the bottom of every image, we visualize the corresponding attention distributions produced by the model.
As we can see from the figure, the heatmaps are not conclusive. They are diffuse and does not clearly show one object where the model pays attention to.

\section{Robustification by Data Augmentation}

\begin{table} 
\small
\centering
\begin{tabular}{l  c c c  } 
 \toprule
 & C+L (\%) & SAAA (\%) & SNMN (\%) \\ [0.3ex]
 \midrule
 Accuracy orig & 60.21 & 70.26 & 66.04 \\

 \midrule
 Predictions flipped & 17.15 & 7.53 & 6.38\\
 neg$\rightarrow$pos &  6.79 & 2.71 & 2.54 \\
 pos$\rightarrow$neg & 7.34 & 3.42 & 2.84  \\
 neg$\rightarrow$neg & 3.02 & 1.39 & 1.01 \\
 \bottomrule
\end{tabular}
\caption{ Accuracy-flipping on VQA-IR edit test split with zero overlap.}
\label{table:flipping_0.1_0.0}
\end{table}
\subsection{Models Performance}
For our fine-tuning experiments, we use a strict subset of \vqaiv with an overlap score of zero. As promised in the paper, Table \ref{table:flipping_0.1_0.0} shows the accuracy-flipping analysis for all the models on this strict subset.  As we see, the numbers are comparable to the model's performance on the overall set.

\subsection{InVariant VQA Augmentation}

\begin{figure} 
\centering
\small
  \begin{tabular}{l  c c c c}

    \toprule

    \multicolumn{5}{c}{Q: What color is the floor? } \\
    \multicolumn{3}{c}{\myit{A}: green} & \multicolumn{2}{c}{\textit{chair removed}; \myit{A}: green} \\
    \multicolumn{5}{c}{\includegraphics[width=0.48\textwidth, trim= 0 0 0 0, clip]{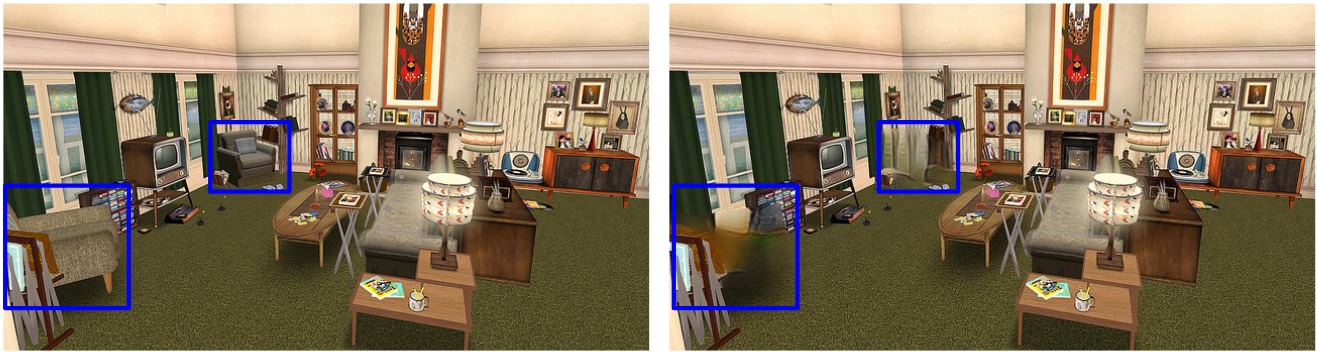}}\\
    & real & real+edit & \tab[1.3cm] real & real+edit\\ 
    CL \tab & \red{brown} & \red{brown}  & \tab[1.3cm] \red{white} &\red{brown} \\
    SAAA  & \green{green} & \green{green}  & \tab[1.3cm] \red{gray} &\green{green}  \\ 
    SNMN &  \green{green} & \green{green}  & \tab[1.3cm] \green{green} &\green{green}  \\

     \toprule

    \multicolumn{5}{c}{Q: Is this a bookstore? } \\
    \multicolumn{3}{c}{\myit{A}: no} & \multicolumn{2}{c}{\textit{person removed}; \myit{A}: no} \\
    \multicolumn{5}{c}{\includegraphics[width=0.48\textwidth, trim= 0 40 0 20, clip]{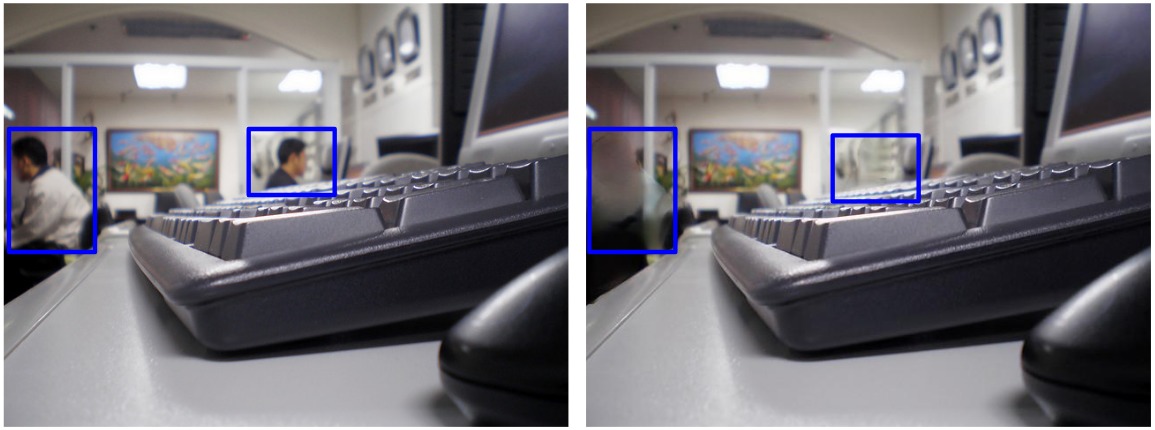}}\\
    & real & real+edit & \tab[1.3cm] real & real+edit\\ 
    CL & \red{yes} & \green{no}  & \tab[1.3cm] \red{yes} &\green{no} \\
    SAAA  & \green{no} & \green{no}  & \tab[1.3cm] \green{no} &\green{no}  \\ 
    SNMN &  \green{no} & \green{no}  & \tab[1.3cm] \red{yes} &\green{no}  \\
     \toprule

    \multicolumn{5}{c}{Q: Is there a pier in the picture? } \\
    \multicolumn{3}{c}{\myit{A}: yes} & \multicolumn{2}{c}{\textit{boat removed}; \myit{A}: yes} \\
    \multicolumn{5}{c}{\includegraphics[width=0.48\textwidth, trim= 0 40 0 20, clip]{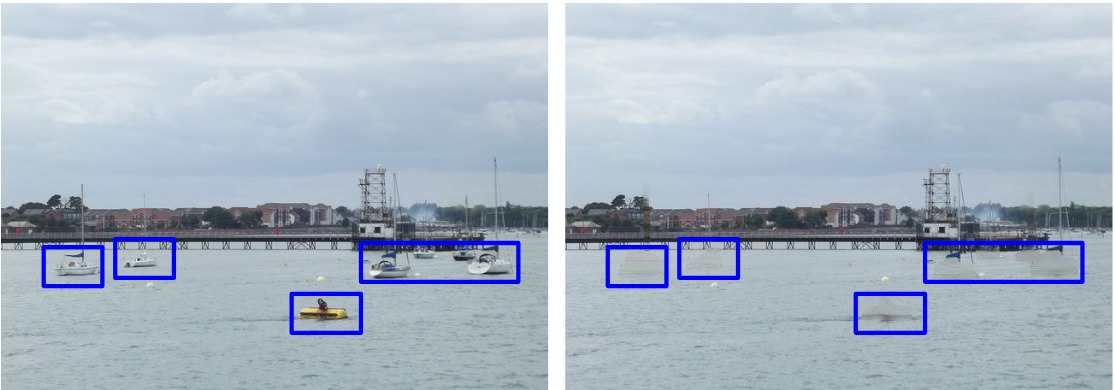}}\\
    & real & real+edit & \tab[1.3cm] real & real+edit\\ 
    CL & \green{yes} & \green{yes}  & \tab[1.3cm] \green{yes} &\green{yes} \\
    SAAA  & \red{no} & \green{yes}  & \tab[1.3cm] \red{no} &\green{yes}  \\ 
    SNMN &  \green{yes} & \green{yes}  & \tab[1.3cm] \red{no} &\green{no}  \\
     \toprule

    \multicolumn{5}{c}{Q: How many bowls of food are there? } \\
    \multicolumn{3}{c}{\myit{A}: 2} & \multicolumn{2}{c}{\textit{bottle removed}; \myit{A}: 2} \\
    \multicolumn{5}{c}{\includegraphics[width=0.48\textwidth, trim= 0 40 0 20, clip]{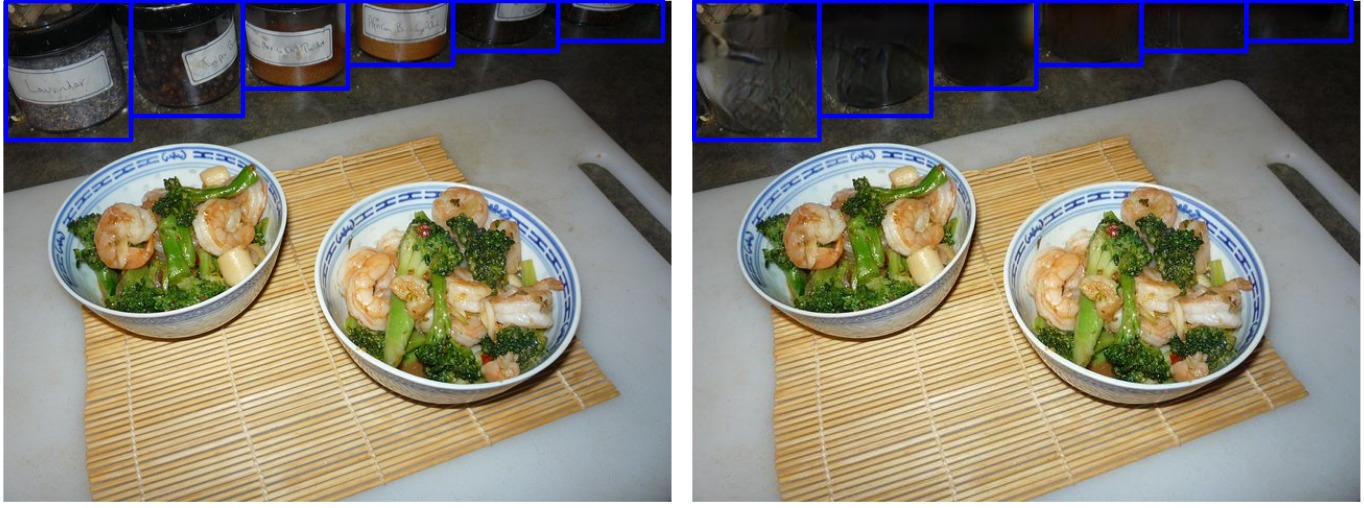}}\\
    & real & real+edit & \tab[1.3cm] real & real+edit\\ 
    CL & \green{2} & \green{2}  & \tab[1cm] \red{3} &\green{2} \\
    SAAA  & \green{2} & \green{2}  & \tab[1cm] \green{2} &\green{2}  \\ 
    SNMN &  \green{2} & \green{2}  & \tab[1cm] \red{1} &\green{2}  \\
    \toprule

  \end{tabular}
  \caption{InVariant VQA Augmentation: Some visualizations from fine-tuning experiments using real/real+edit data from \vqaiv. Using real+edit makes models more consistent.}
  \label{fig:DA_visual_aid_IV}
\end{figure}

\begin{figure} 
\centering
\small
  \begin{tabular}{l  c c c c}

    \toprule
    \multicolumn{5}{c}{Q: How many planes are in the air? } \\
    \multicolumn{3}{c}{\myit{A}: 1} & \multicolumn{2}{c}{\textit{plane removed}; \myit{A}: 0} \\
    \multicolumn{5}{c}{\includegraphics[width=0.48\textwidth, trim= 0 0 0 0, clip]{Figures/planes_1}}\\
    & real & real+edit & \tab[1.3cm] real & real+edit\\ 
    CL & \green{1} & \green{1}  &  \tab[1.3cm] \red{1} &\green{0} \\
    SAAA  & \green{1} & \green{1}  & \tab[1.3cm] \red{1} &\green{0}  \\ 
    SNMN &  \green{1} & \green{1}  & \tab[1.3cm] \red{1} &\green{0}  \\
    
    \toprule

    \multicolumn{5}{c}{Q: How many zebras are there in the picture? } \\
    \multicolumn{3}{c}{\myit{A}: 2} & \multicolumn{2}{c}{\textit{zebra removed}; \myit{A}: 1} \\
    \multicolumn{5}{c}{\includegraphics[width=0.48\textwidth, trim= 0 0 0 0, clip]{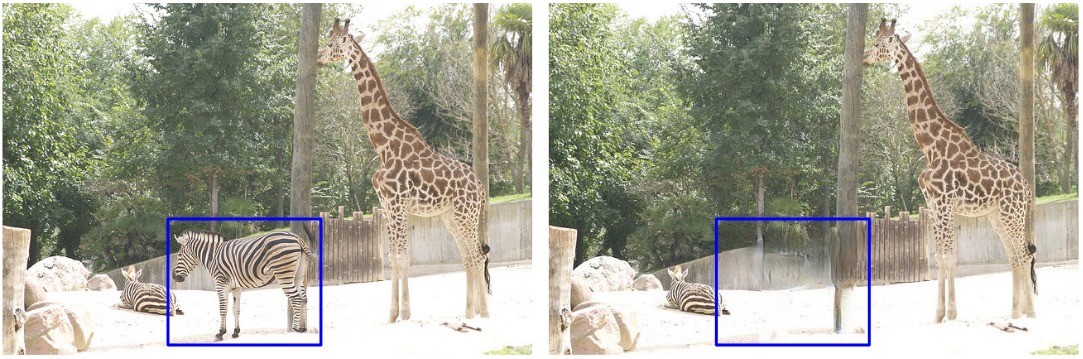}}\\
    & real & real+edit & \tab[1.3cm] real & real+edit\\ 
    CL & \green{2} & \green{2}  & \tab[1.3cm] \red{2} &\green{1} \\
    SAAA  & \green{2} & \green{2}  & \tab[1.3cm] \red{2} &\green{1}  \\ 
    SNMN &  \green{2} & \green{2}  & \tab[1.3cm] \red{2} &\green{1}  \\
    
    \toprule
    
    \multicolumn{5}{c}{Q: How many boys are playing Frisbee? } \\
    \multicolumn{3}{c}{\myit{A}: 2} & \multicolumn{2}{c}{\textit{person removed}; \myit{A}: 1} \\
    \multicolumn{5}{c}{\includegraphics[width=0.48\textwidth, trim= 0 0 0 0, clip]{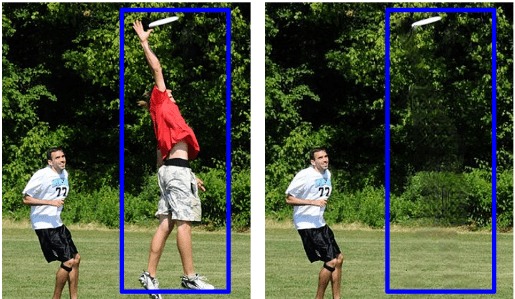}}\\
    & real & real+edit & \tab[1.3cm] real & real+edit\\ 
    CL & \red{1} & \green{2}  & \tab[1.3cm] \green{1} &\green{1} \\
    SAAA  & \green{2} & \green{2}  & \tab[1.3cm] \red{2} &\green{1}  \\ 
    SNMN &  \red{1} & \green{2}  & \tab[1.3cm] \green{1} &\green{1}  \\
    
    \toprule
    
  \end{tabular}
  \caption{CoVariant VQA Augmentation: Some visualizations from fine-tuning experiments using real/real+edit data from both \vqacv and \vqaiv. Using real+edit makes models more consistent and in these examples- also accurate.}
  \label{fig:DA_visual_aid_CV}
\end{figure}

In Table \ref{table:DA_ques_types}, we show the accuracy-flipping analysis for all the specialized models. Figure \ref{fig:DA_visual_aid_IV} shows some examples where using additional synthetic data makes models more consistent. In the paper, we show a compact representation of the table by plotting the reduction in flips/improvement in accuracy for models finetuned using real+edit data relative to the models finetuned only using the real data. For instance, for question-type `is this a' for CNN+LSTM: we see there is (12.72-9.77)/12.72 which is about 23\% reduction in flips. These numbers show that using synthetic data always leads to a reduction in flips and in some cases- also results in improved accuracy on the original VQA set.
Figure \ref{fig:DA_visual_aid_CV} shows some of the examples where using the synthetic data makes the models n/n-1 consistent and accurate as well.

\begin{table} 
\small
\centering
\begin{tabular}{l  l l l}

 \toprule
 
 & CL (\%) & SAAA (\%) & SNMN (\%)  \\ 
 \midrule
 \multicolumn{4}{c}{what color is the} \\
 Acc orig & 65.48 \arr 65.06  & 82.12\arr83.75  & 78.78\arr80.1\\   
 Pred flipped & 11.79 \arr 10.89  & 7.25\arr6.27 & 7.41\arr7.21\\

 \midrule
 
 \multicolumn{4}{c}{is there a } \\
 Acc orig & 61.96\arr63.61 & 69.44\arr69.36 &  71.32\arr72.26\\
 Pred flipped & 13.3\arr10.81 & 8.75\arr7.51 &  8.83\arr7.79\\

 \midrule

 \multicolumn{4}{c}{is this a} \\
 Acc orig & 64.99\arr64.87 &  74.33\arr72.84  & 76.54\arr76.79 \\
 Pred flipped  & 12.72\arr9.77  & 6.09\arr5.14  &  6.96\arr6.52 \\

 \midrule

 \multicolumn{4}{c}{how many }\\
 Acc orig & 43.24\arr43.2 &  50.38\arr50.12  & 49.71\arr50.56 \\
 Pred flipped  & 21\arr20.1  & 13.35\arr11.04  &  14.04\arr13.35 \\
 \midrule

 \multicolumn{4}{c}{counting} \\
 Acc orig &42.87\arr43.58 &  51.05\arr49.94 &  50.22\arr50.26 \\
 Pred flipped & 21.08\arr19.06 & 12.81\arr12.60  & 14.76\arr12.96 \\
 \bottomrule

\end{tabular}
\caption{ \vqaiv Augmentation: numbers on the left side of the arrow denote the accuracy/flipping for models fine-tuned using just real data whereas numbers on the right side show the performance of models when finetuned with real+synthetic data}
\label{table:DA_ques_types}
\end{table}

\subsection{CoVariant VQA Augmentation}

Table \ref{table:count_del1_DA} shows the accuracy/flips for models fine-tuned using real/ real+synthetic IQAs. In the paper we compress the information in the form of plot as we do in the case of \vqaiv augmentation.

\begin{table} 
\small
\centering
\begin{tabular}{l  l l l} 
 \toprule
 & CL (\%) & SAAA (\%) & SNMN (\%)  \\ 
 \midrule
 \multicolumn{4}{c}{\vqacv }\\
 Acc orig & 43.65\arr42.04  &   50.87\arr50.24  & 50.67\arr49.99 \\   
 Pred flipped & 83.84 \arr 50.74 &  77.74\arr45.85  &73.12\arr44.19  \\ [0.5ex]

\midrule
 \multicolumn{4}{c}{\vqacv+\vqaiv }\\
 Acc orig & 43.65\arr43.94  &      50.87\arr50.45   & 50.67\arr50.61\\   
 Pred flipped & 83.84  \arr59.58 &  77.74\arr52.71  & 73.12\arr51.91 \\

 \bottomrule
\end{tabular}
\caption{\vqacv Augmentation: numbers on the left side of the arrow denote the accuracy/flipping for models fine-tuned using just real data whereas numbers on the right side show the performance of models when finetuned with real+synthetic data}
\label{table:count_del1_DA}
\end{table}

\section{Outlook on building causal VQA models}
In recent works~\cite{arjovsky2019invariant, heinze2018conditional}, image classifiers are taught to rely on causal features by imposing regularization across data from different environments/ identities. A requirement for this is to have pairs of data points where the only change is in non-causal features, so one can regularize the network response to these. In our work, we explicitly create such data for the VQA task. We believe, future work can exploit this data for imposing consistency losses across original/edited IQA triplets while training or providing part of the causal structure as part of the supervision.

\FloatBarrier
\end{appendices}

\end{document}